\documentclass[11pt,letterpaper]{article}
\usepackage[T1]{fontenc}
\usepackage[utf8]{inputenc}
\usepackage{newtxtext,newtxmath}
\usepackage[margin=0.86in,top=0.78in,bottom=0.82in,headheight=24pt,headsep=16pt]{geometry}
\usepackage{microtype}
\usepackage[table]{xcolor}
\usepackage{graphicx}
\usepackage{booktabs,tabularx,array}
\usepackage{amsmath}
\usepackage{enumitem}
\usepackage{titlesec}
\usepackage{fancyhdr}
\usepackage{caption,xcolor}
\usepackage[numbers,sort&compress]{natbib}
\setcitestyle{square,comma}
\usepackage{tikz}
\usetikzlibrary{arrows.meta,positioning,calc,fit}
\usepackage{hyperref}

\definecolor{Ink}{HTML}{22252B}
\definecolor{Muted}{HTML}{627078}
\definecolor{Slate}{HTML}{263B66}
\definecolor{Indigo}{HTML}{4362C7}
\definecolor{Teal}{HTML}{195E69}
\definecolor{Purple}{HTML}{995692}
\definecolor{Sand}{HTML}{64748B}
\colorlet{FamilyRose}{Purple}
\definecolor{SlateTint}{HTML}{F5F6F8}
\definecolor{NeutralTint}{HTML}{F3F4F6}
\definecolor{Rule}{HTML}{DCE1E7}
\definecolor{SharedFill}{HTML}{DCE5EE}
\definecolor{BranchFill}{HTML}{E8E5EF}
\definecolor{FusionFill}{HTML}{E9E7EC}
\colorlet{MBTBlue}{Indigo}
\colorlet{MBTTeal}{Teal}
\colorlet{MBTPurple}{Purple}
\colorlet{MBTTwo}{MBTBlue}
\colorlet{MBTThree}{MBTTeal}
\colorlet{MBTFour}{MBTPurple}
\colorlet{MBTMoEPurple}{Purple}
\colorlet{SDInk}{Ink}
\colorlet{SDSlate}{Slate}
\colorlet{SDBlue}{MBTBlue}
\colorlet{SDTeal}{MBTTeal}
\colorlet{SDPurple}{MBTPurple}
\colorlet{SDNeutral}{Rule}

\DeclareRobustCommand{\mbt}{MBT}
\newcommand{\ReportTitle}{Multi-Branch Transformer}
\newcommand{\ReportAuthors}{Gaotang Li, Lingjie Chen, Chunyuan Deng, Zhiwei Xu, Hongzheng Chen, Christopher De Sa, Jingrui He, Yingheng Wang, Hanghang Tong}

\hypersetup{
  colorlinks=true,linkcolor=Slate,citecolor=Slate,urlcolor=Slate,
  pdftitle={\ReportTitle},
  pdfauthor={\ReportAuthors},
  pdfsubject={Multi-Branch Transformer report template},
  bookmarksnumbered=true
}
\color{Ink}
\setlist[itemize]{leftmargin=1.25em,itemsep=2pt,topsep=3pt}
\titleformat{\section}{\large\bfseries\color{Slate}}{\thesection.}{0.55em}{}
\titleformat{\subsection}{\normalsize\bfseries\color{Ink}}{\thesubsection}{0.55em}{}
\titleformat{\subsubsection}{\normalsize\bfseries\color{Ink}}{\thesubsubsection}{0.55em}{}
\titlespacing*{\subsubsection}{0pt}{8pt plus 1pt}{3pt}
\titlespacing*{\section}{0pt}{15pt plus 2pt minus 2pt}{5pt}
\titlespacing*{\subsection}{0pt}{9pt plus 1pt}{3pt}
\renewcommand{\arraystretch}{1.22}
\newcolumntype{Y}{>{\raggedright\arraybackslash}X}
\newcolumntype{Z}{>{\centering\arraybackslash}X}

\renewcommand{\headrulewidth}{0.35pt}
\renewcommand{\headrule}{\hbox to\headwidth{\color{Rule}\leaders\hrule height \headrulewidth\hfill}}
\newcommand{\MBTEmailLine}{\fontsize{5.8}{8}\selectfont\color{Muted}\mbox{gaotang3@illinois.edu; lingjie7@illinois.edu; chunyuan.deng@rice.edu; zhiweixu@umich.edu; hc676@cornell.edu; cmd353@cornell.edu; jingrui@illinois.edu; yingheng@cs.cornell.edu; htong@illinois.edu}}

\newsavebox{\MBTEmailBox}
\fancypagestyle{firstpage}{%
  \fancyhf{}%
  \fancyfoot[L]{\fontsize{6.4}{8}\selectfont\color{Muted}%
    \sbox{\MBTEmailBox}{\MBTEmailLine}%
    \raisebox{-\dp\MBTEmailBox}[\ht\MBTEmailBox][\dp\MBTEmailBox]{%
      \includegraphics[width=\wd\MBTEmailBox,totalheight=\dimexpr\ht\MBTEmailBox+\dp\MBTEmailBox\relax]{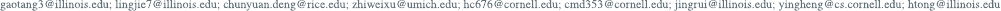}}}%
  \renewcommand{\headrulewidth}{0pt}%
  \renewcommand{\footrule}{\hbox to\headwidth{\color{Slate!55}\leaders\hrule height 0.4pt\hfill}}%
}

\begin{document}

\thispagestyle{firstpage}

{\color{Slate!65}\rule{\linewidth}{0.7pt}}\par
\vspace{6pt}
{\fontsize{30.5}{35}\selectfont\bfseries\color{Slate}%
\raisebox{-2.9pt}[0pt][0pt]{\includegraphics[width=26pt]{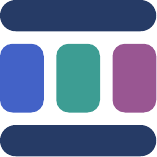}}\hspace{10.2pt}%
\ReportTitle\par}
\vspace{12pt}
\begingroup
\setlength{\parskip}{0pt}
\newcommand{\MBTMarkSize}{5.6}
\newcommand{\MBTMarkRaise}{2.8pt}
\newcommand{\MBTAuthorMark}[1]{%
  \raisebox{\MBTMarkRaise}{\normalfont\fontsize{\MBTMarkSize}{\MBTMarkSize}\selectfont #1}}
{\renewcommand{\MBTMarkSize}{6.65}\renewcommand{\MBTMarkRaise}{3.3pt}%
\fontsize{9.75}{12.1}\selectfont\bfseries\raggedright
\mbox{Gaotang~Li\MBTAuthorMark{1,*,\textdagger}},\hspace{.20em}%
\mbox{Lingjie~Chen\MBTAuthorMark{1,\textdagger}},\hspace{.20em}%
\mbox{Chunyuan~Deng\MBTAuthorMark{2,\textdagger}},\hspace{.20em}%
\mbox{Zhiwei~Xu\MBTAuthorMark{3,\textdagger}},\hspace{.20em}%
\mbox{Hongzheng~Chen\MBTAuthorMark{4,\textdagger}},\hspace{.20em}%
\mbox{Christopher~De~Sa\MBTAuthorMark{4}},\\
\mbox{Jingrui~He\MBTAuthorMark{1}},\hspace{.20em}%
\mbox{Yingheng~Wang\MBTAuthorMark{4,\textdagger,\textsection}},\hspace{.20em}%
\mbox{Hanghang~Tong\MBTAuthorMark{1,\textsection}}\par}
\vspace{3pt}
{\renewcommand{\MBTMarkSize}{6.35}\renewcommand{\MBTMarkRaise}{3.17pt}%
\fontsize{8.5}{10.2}\selectfont
\MBTAuthorMark{1}\,University of Illinois Urbana-Champaign\quad
\MBTAuthorMark{2}\,Rice University\quad
\MBTAuthorMark{3}\,University of Michigan\quad
\MBTAuthorMark{4}\,Cornell University\par}
\vspace{3pt}
{\renewcommand{\MBTMarkSize}{5.6}\renewcommand{\MBTMarkRaise}{2.8pt}%
\fontsize{7.5}{9}\selectfont\color{Muted}
\MBTAuthorMark{*}\,Project lead\qquad
\MBTAuthorMark{\textdagger}\,Core contributors\qquad
\MBTAuthorMark{\textsection}\,Last authors\par}
\vspace{8pt}
{\color{Rule}\hrule height 0.4pt}
\endgroup
\vspace{6pt}

\begingroup
\setlength{\fboxsep}{7pt}
\colorbox{SlateTint}{%
  \begin{minipage}{\dimexpr\linewidth-2\fboxsep\relax}
    {\fontsize{12}{14}\selectfont\bfseries\color{Slate}Abstract}\par\vspace{3pt}
    \fontsize{10}{12.2}\selectfont

Transformer language models are commonly scaled by making their blocks wider or stacking them deeper.
In this work, we explore a complementary path: \textbf{branching}, the parallel composition of full Transformer blocks.
We introduce the \textbf{Multi-Branch Transformer} (\textbf{\mbt{}}), which processes the input through shared initial layers, executes independently parameterized branches of sequential blocks in parallel, and fuses their outputs before passing the result through shared final layers.
At comparable training compute relative to a standard transformer, we study three \mbt{} variants: Wide-Shallow preserves block width, Narrow-Deep preserves sequential depth, and Balanced interpolates between these designs.
Across controlled pretraining experiments at four mixture-of-experts model scales, ranging from 0.8B to 2.1B total parameters, we show that 2-branch Wide-Shallow \mbt{}s consistently outperform sequential Transformers under matched parameter and training-compute budgets.
Balanced \mbt{}s further improve modeling performance over their Wide-Shallow counterparts.
These models retain comparable quality as the number of branches increases from 2 to 3 or 4.
To further improve inference efficiency at long contexts, we introduce \textbf{Joint-KV}, which constructs a shared KV cache from branch representations at each depth.
Our final model recipe combines Balanced \mbt{}s with Joint-KV, supported by an inference engine that exploits branch grouping and shared-cache reuse.
Fitted scaling curves indicate that standard Transformers require up to \textbf{1.26$\times$} as much training body compute to match our final models' validation loss.
Across evaluated configurations, these models reduce full-model KV-cache storage by up to \textbf{34\%} and deliver up to \textbf{51.4\%} higher decoding throughput on a single H200 GPU.
Extensive ablation studies show that branching yields gains beyond changes in width and depth, enables an effective KV-sharing strategy (Joint-KV), remains robust across fusion functions, and performs best in the middle layers.
Together, these findings demonstrate that branching offers a practical path for scaling language models toward both better modeling quality and faster inference.
\par\vspace{4pt}
{\fontsize{10}{12.2}\selectfont
\raisebox{-1.6pt}{\includegraphics[height=11pt]{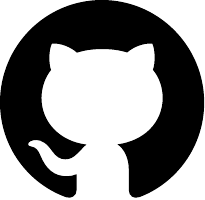}}\hspace{.35em}%
\textbf{Code:} \href{https://github.com/GaotangLi/Multi-Branch-Transformer}{\nolinkurl{https://github.com/GaotangLi/Multi-Branch-Transformer}}\\[1pt]
\raisebox{-2pt}{\includegraphics[height=12.5pt]{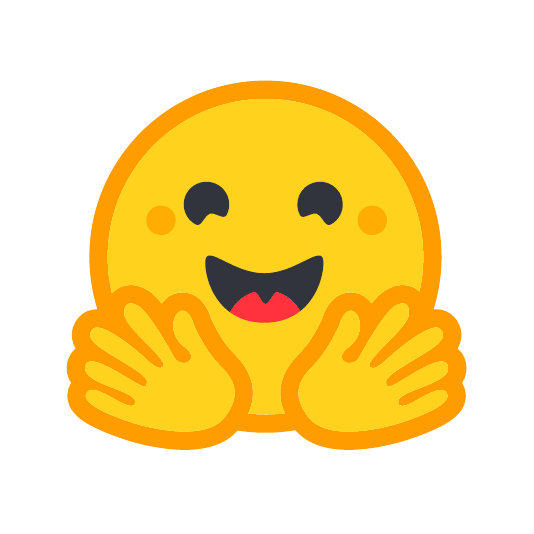}}\hspace{.35em}%
\textbf{Hugging Face:} \href{https://huggingface.co/collections/gaotang/multi-branch-transformer}{\nolinkurl{https://huggingface.co/collections/gaotang/multi-branch-transformer}}\par}

  \end{minipage}%
}\par
\endgroup

\enlargethispage{12pt}
\par\vspace*{\fill}
\noindent\begin{minipage}{\linewidth}
  \setlength{\parskip}{0pt}
  \captionsetup{type=figure,skip=6pt}
  \centering
  \makebox[\linewidth][c]{\includegraphics{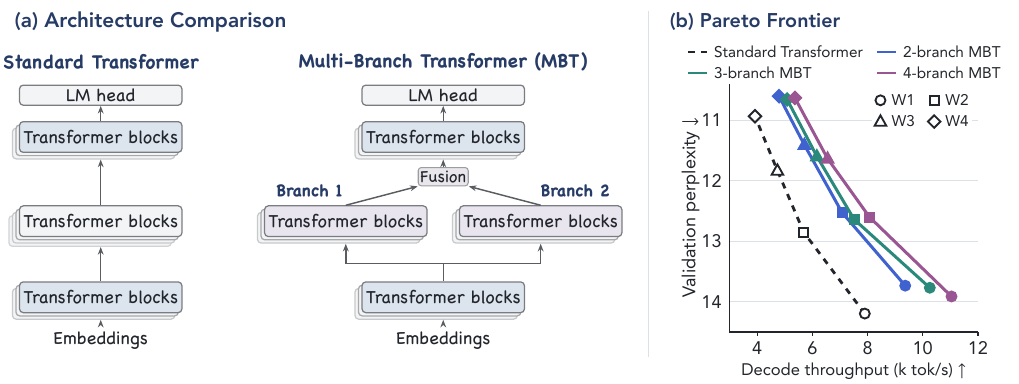}}%
  \caption{(a) 
  Architecture comparison between a standard transformer and a Multi-Branch Transformer (\mbt{}):
  MBT composes full blocks in parallel branches. (b) Validation perplexity versus decoding throughput for the Standard Transformer and 2-, 3-, and 4-branch MBTs on a single H200 GPU (16K input tokens, 64 requests). The y-axis is reversed, with the upper right indicating better quality and higher throughput. W1--W4 denote the four model scales in our scaling ladder.}
  \label{fig:overview}
\end{minipage}\par
\vspace*{\fill}

\clearpage

\section{Introduction}
\label{sec:introduction}

Scaling Transformer~\citep{vaswani2017attention} language models involves deciding both how much parameter capacity to allocate and how to organize it~\citep{kaplan2020scaling,hoffmann2022training}.
Depth and width are two conventional architectural choices: stacking more blocks in sequence or enlarging the transformations within each block~\citep{tay2022scale,mcleish2025gemstones,bian2025scaling}.
Recent advances in LLM architectures have substantially expanded the design space within this template through innovations in attention mechanisms~\citep{kimi2025linear,gemma2025gemma3}, sparse expert computation~\citep{deepseek2025v3}, and residual connections~\citep{zhu2024hyperconnections,xie2025mhc,kimi2026attention}, while the \emph{sequential composition} of full blocks remains a common foundation.
In this work, we explore \textbf{branching}, the parallel composition of full Transformer blocks, as a complementary way to organize and scale model parameters, and ask:

\begin{center}
    \emph{Can Transformers scale more effectively and efficiently through parallel compositions of full blocks?}
\end{center}

\begin{figure}[tbp!]
  \centering

\input{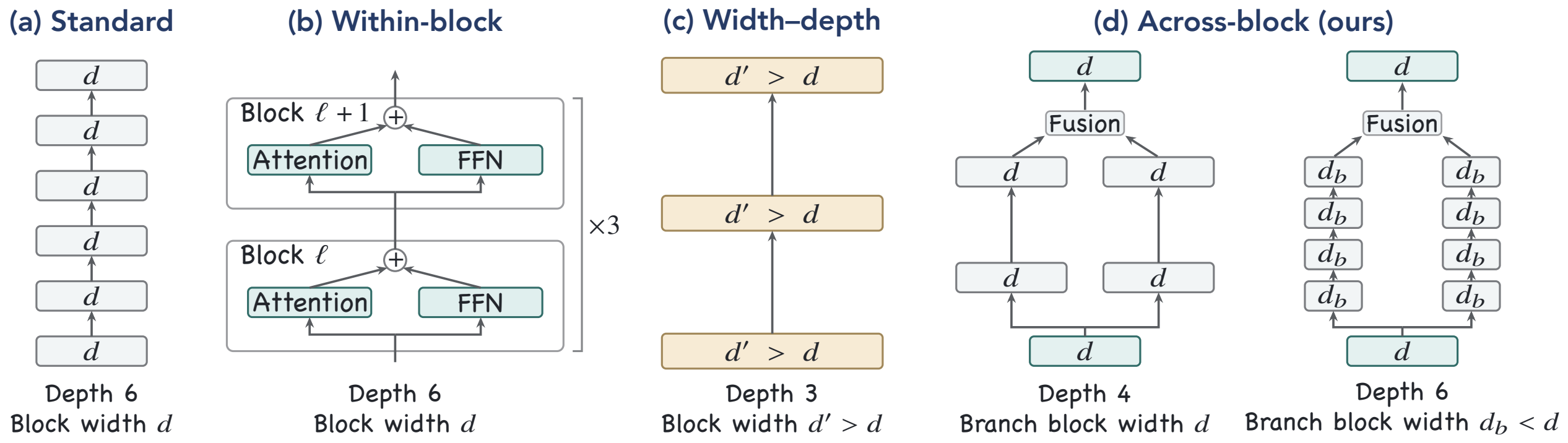}%

  \caption{\textbf{Comparison of Transformer computation structures.}
(a) Standard sequential blocks; (b) within-block parallel attention and FFN, as in PaLM~\citep{chowdhery2022palm}; (c) wider, shallower stacks; and (d) \mbt{}s with concurrent, independently parameterized branches of full Transformer blocks.
Depth denotes the number of sequential blocks along a computation path, excluding fusion.
\mbt{}s explore an orthogonal axis of parallelism at the granularity of full Transformer blocks.}
  \label{fig:width-branch-motivation}
\end{figure}

To this end, we introduce the \textbf{Multi-Branch Transformer} (\textbf{\mbt{}}), which enables parallel computation at the granularity of full Transformer blocks (Figure~\ref{fig:overview}(a)). An \mbt{} processes the input through shared initial layers (the \texttt{prelude}\footnote{We use the terminology \texttt{prelude} and \texttt{coda} following \citet{geiping2025scaling}.})
, splits computation across $K$ independently parameterized branches that execute concurrently, fuses their outputs, and passes the result through shared final layers (the \texttt{coda}).
The key idea is to preserve sequential composition within each branch while allowing subsequent shared layers to build on the results of multiple transformation paths.
In comparison, related work on parallel Transformer architectures explores different design axes (Figure~\ref{fig:width-branch-motivation}): concurrency within individual blocks~\citep{chowdhery2022palm}, allocation of width and depth~\citep{tay2022scale,mcleish2025gemstones,bian2025scaling}, or alternative parallel computation paths~\citep{chen2025parallel}.
Systems approaches such as tensor and pipeline parallelism~\citep{narayanan2021efficient} are orthogonal to and compatible with this architectural design.

Relative to a standard Transformer, an \mbt{} can be defined in three ways at comparable compute: Wide-Shallow \mbt{}s preserve the baseline’s block width and distribute its blocks across branches; Narrow-Deep \mbt{}s preserve the baseline’s depth along each branch using narrower blocks; and Balanced \mbt{}s interpolate between these choices by jointly adjusting branch width and depth.

To further exploit the parallel branch structure of \mbt{}s, we propose \textbf{Joint-KV}, which constructs a shared key-value (KV) cache from all branch representations at each depth while retaining separate queries for each branch (Figure~\ref{fig:kv-sharing}).
For $K$ branches, this reduces cache storage in the branch region to $1/K$ of Independent-KV at fixed branch depth and KV dimensions.
We combine Joint-KV with Balanced \mbt{}s as our final model recipes (Figure~\ref{fig:overview}(b)).
To translate our architectural design into practical inference gains, we develop an inference engine based on nano-vLLM~\citep{kwon2023pagedattention,yu2025nanovllm} that exploits the parallel branches and shared KV caches through \textbf{branch grouping} and head folding, reducing sequential execution steps and redundant KV reads during decoding.

Extensive experiments validate the effectiveness and efficiency of \mbt{}.
First, branching improves \textbf{language modeling performance}.
Across a scaling ladder of mixture-of-experts models (0.8B-2.1B total parameters), our scaling laws first indicate that the best 2-branch Wide-Shallow \mbt{} matches a standard Transformer trained with \textbf{1.23$\times$} as much compute, while allocating approximately 79\% of blocks to parallel branches yields performance on par with the standard Transformer. 
Balanced \mbt{}s achieve the strongest performance among the studied allocations and maintain comparable quality as the number of branches increases.
Second, Joint-KV \mbt{}s, together with our inference engine, substantially improve \textbf{inference speed} while retaining strong modeling performance.
In particular, 2-, 3-, and 4-branch Joint-KV \mbt{}s match standard Transformers trained with \textbf{1.16--1.26$\times$} as much training body compute, while reducing KV cache size by \textbf{19--34\%}.
On a single H200 GPU, our models achieve up to \textbf{51.4\%} higher decoding throughput than the standard Transformer with 64K-token inputs and 8 concurrent requests.

Extensive ablation studies spanning five scaling sweeps further demonstrate that the benefits of branching extend beyond conventional adjustments to model width and depth. 
At matched KV-cache capacity, sharing KV across branches (Joint-KV) achieves better language modeling performance than keeping a private KV cache in each branch, highlighting the benefit of exploiting \mbt{}'s branch structure for improving long-context inference efficiency.
We also find that \mbt{}s are robust to the choice of fusion function and perform best when branches are placed in the middle layers.
Together, these findings highlight the organization of full Transformer blocks as a promising architectural dimension for improving both language modeling quality and inference efficiency.

\setcounter{topnumber}{1}
\section{Multi-Branch Transformer}
\label{sec:method}

An \mbt{} replaces part of a serial Transformer stack with independently
parameterized branches (Figure~\ref{fig:overview}(a)). Shared blocks
precede the branches and follow their fusion; we call these stacks the
\texttt{prelude} and \texttt{coda}, respectively.

\subsection{Architecture}
\label{sec:method-architecture}

Let $x=(x_1,\ldots,x_T)$ be an input token sequence and
$H\in\mathbb{R}^{T\times d}$ its hidden states at a block's input,
where $d$ is the residual width. Each block $\mathcal{T}_\ell$ applies causal attention followed by a
mixture-of-experts (MoE) feed-forward module~\citep{fedus2022switch}:
\begin{equation}
  H'=H+\operatorname{Attn}_\ell(H),\qquad
  \mathcal{T}_\ell(H)=H'+\operatorname{MoE}_\ell(H').
  \label{eq:transformer-block}
\end{equation}
Here $H'$ is the hidden state after the attention residual update;
the blocks' pre-normalization is omitted for clarity. A standard Transformer
embeds the tokens and composes $L$ blocks before decoding:
\begin{equation}
  X=\operatorname{Emb}(x),\qquad
  \widehat{Y}=\operatorname{Dec}\bigl((\mathcal{T}_L\circ\cdots\circ\mathcal{T}_1)(X)\bigr).
  \label{eq:sequential-blocks}
\end{equation}
The decoder $\operatorname{Dec}$ applies final normalization, the LM head,
and softmax; $\widehat{Y}_t$ is the predicted distribution of $x_{t+1}$.

An \mbt{} replaces the serial middle stack with $K\geq2$ independently
parameterized branch stacks $\mathcal{G}_i$. The \texttt{prelude} $\mathcal{P}$,
fusion $\mathcal{F}$, and \texttt{coda} $\mathcal{C}$ connect them:
\begin{equation}
  H=\mathcal{P}(X),\quad
  Z_i=\mathcal{G}_i(H),\quad
  \widehat{Y}=\operatorname{Dec}\bigl(\mathcal{C}(\mathcal{F}(Z_1,\ldots,Z_K))\bigr).
  \label{eq:branch-organization}
\end{equation}
Here $H$ is the shared \texttt{prelude} output, and $\mathcal{G}_i$ is a stack of
$L_i$ blocks ($i=1,\ldots,K$). When widths differ, $\mathcal{P}$ includes a
shared input projection to the branch width.

\noindent\textbf{Fusion.}
At each token position, we normalize the branch outputs, concatenate them along the hidden dimension, and project the result back to the shared width $d$:
\begin{equation}
  \mathcal{F}(Z_1,\ldots,Z_K)
  =\operatorname{Concat}\bigl(\mathcal{N}_1(Z_1),\ldots,\mathcal{N}_K(Z_K)\bigr)W_F.
  \label{eq:branch-fusion}
\end{equation}
Each $\mathcal{N}_i$ is an RMSNorm, and $W_F$ is a bias-free linear map.
The projected output goes directly to the \texttt{coda}, with no separate branch
residual.
We explore more expressive fusion in  Section~\ref{sec:analysis-fusion}, but find Eq.~\ref{eq:branch-fusion} to be sufficient.

\begin{figure}[t]
  \centering
  \input{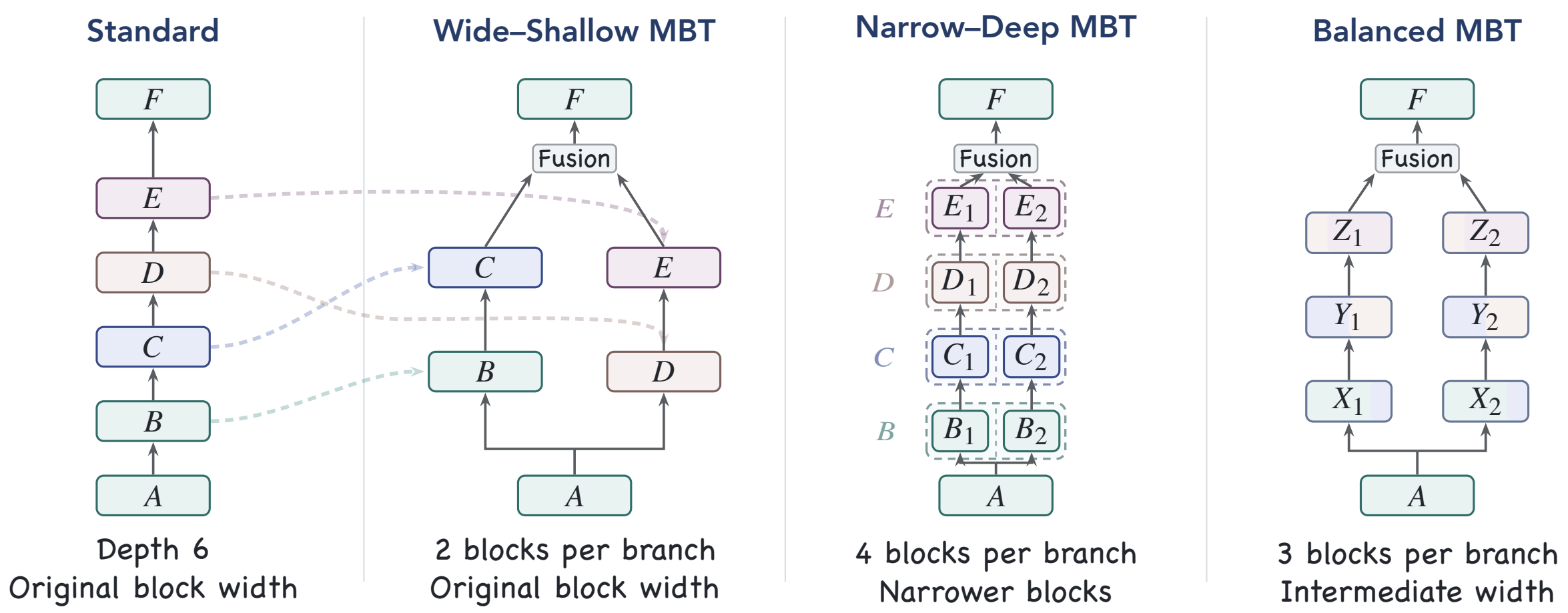}%
  \vspace{+0.4em}
  \caption{\textbf{Three width--depth allocations for \mbt{}s.} Relative to a standard Transformer under matched compute, (1) Wide-Shallow \mbt{} preserves block width and redistributes the middle blocks across branches; (2) Narrow-Deep \mbt{} preserves the depth along each branch using narrower blocks; and (3) Balanced \mbt{} uses intermediate width and depth.}
  \label{fig:mbt-families}
\end{figure}

\subsection{Width--Depth Allocation}
\label{sec:method-families}

Let $P$ and $C$ denote the numbers of \texttt{prelude} and \texttt{coda} blocks
($L_{\rm in}$ and $L_{\rm out}$ in Figure~\ref{fig:overview}(a)), and
$L_i$ denote the number of blocks in branch $i$. The total block count $L_{\mathrm{physical}}$ and depth $D$ of an \mbt{} are
\begin{equation}
  L_{\mathrm{physical}}=P+\sum_{i=1}^{K}L_i+C,\qquad
  D=P+\max_{i=1,\ldots,K}L_i+C.
  \label{eq:branch-path}
\end{equation}

With the shared \texttt{prelude} and \texttt{coda} fixed, we compare three branch width–depth allocations under matched training compute:

\begin{itemize}[nosep]
  \item \textbf{Wide--Shallow:} preserve the original block width and redistribute the middle blocks across parallel branches.
  \item \textbf{Narrow--Deep:} preserve the original sequential depth along each branch while reducing block width.
  \item \textbf{Balanced:} use intermediate branch width and depth between these two endpoints.
\end{itemize}

For Narrow--Deep and Balanced, we adjust the residual width and MoE
dimensions within each branch to match the training FLOPs of the
Standard reference. All other configurations (\emph{e.g.,} total expert, top-$k$) remain unchanged.

Figure~\ref{fig:mbt-families} illustrates these variants using a six-block Transformer, $A\to B\to C\to D\to E\to F$. Here, $A$ and $F$ serve as the shared \texttt{prelude} and \texttt{coda}, respectively, while the middle stack is reorganized into two parallel branches. Wide--Shallow places two original-width blocks in each branch; Narrow--Deep uses four narrower blocks per branch; and Balanced uses three blocks per branch with an intermediate width.

\begin{figure}[t]
  \centering
  \resizebox{\linewidth}{!}{\input{figures/kv-sharing.tex}}%

  \vspace{8pt}
  \includegraphics[width=\linewidth]{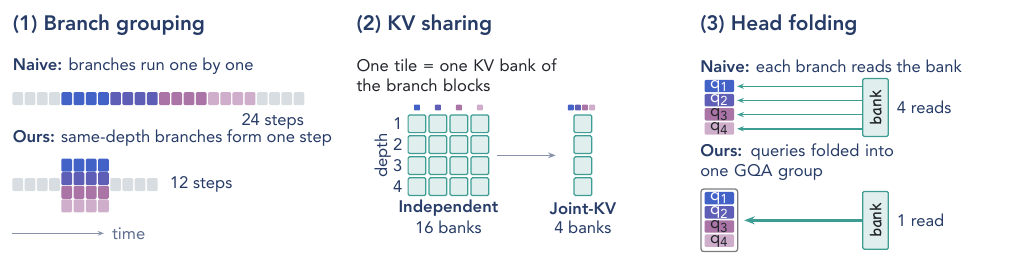}
  \caption{\textbf{KV-cache sharing and serving of \mbt{}s.} Top: KV-cache construction at one branch depth. Independent-KV constructs a separate cache from each branch input.
  Joint-KV projects the concatenated branch inputs into one shared cache,
  while retaining branch-local queries. 
  Bottom: naive versus optimized serving for \mbt{}s (Example in figure: 4-branch \mbt{} at scale W3). (1) Branch grouping executes same-depth blocks in one stage. 
  (2) We call the KV cache of one KV projection a \textit{KV bank}. Independent-KV stores one bank per branch block, and Joint-KV stores one bank per branch depth. 
  (3) All branches read a shared KV bank in one attention call.
  More details are in Figure~\ref{fig:system-counts} in Appendix~\ref{sec:inference-engine}.
  }
  \label{fig:kv-sharing}
  \label{fig:system-design}
\end{figure}

\subsection{KV-Cache Sharing and Infrastructure Design}
\label{sec:method-kv}
\label{sec:system_method}

In the default configuration, which we call Independent-KV, each block in each branch constructs keys and values from its own hidden states and maintains a separate KV cache, so more branches result in more caches. To improve the long-context efficiency of \mbt{}s, we introduce Joint-KV, which shares the KV cache across branches per depth, and an inference engine that turns the parallel branches and the shared caches into faster decoding. Joint-KV changes the model and is used in training. The engine changes only how the model is executed.

\textbf{Joint-KV.}
In Joint-KV, all branches share one KV cache at each depth, constructed jointly from their pre-attention inputs, while queries remain local to each branch. Formally, let $\bar S_{i,\ell}\in\mathbb{R}^{T\times d_b}$ denote the RMS-normalized input to branch $i$ at depth $\ell$. The two KV-cache constructions are
\begin{equation}
  \mathrm{KV}^{\rm ind}_{i,\ell}=\bar S_{i,\ell}W^{\rm ind}_{i,\ell},\qquad
  \mathrm{KV}^{\rm shared}_{\ell}
  =\frac{1}{\sqrt{K}}[\bar S_{1,\ell};\ldots;\bar S_{K,\ell}]W^{\rm shared}_{\ell}.
  \label{eq:kv-construction}
\end{equation}

Let $d_b$ denote the branch width, $h_{\mathrm{KV}}$ the number of KV heads,
and $d_h$ the head dimension. At each depth, Independent-KV uses $K$
learned projection matrices
\mbox{$W^{\rm ind}_{i,\ell}\in\mathbb{R}^{d_b\times (2h_{\mathrm{KV}}d_h)}$},
whereas Joint-KV uses a single
\mbox{$W^{\rm shared}_{\ell}\in\mathbb{R}^{(Kd_b)\times (2h_{\mathrm{KV}}d_h)}$}.
The factor $2$ accounts for keys and values. Both constructions therefore
have the same KV-projection parameter count and matrix-multiplication FLOPs.
$[\cdot;\cdot]$ concatenates along the hidden dimension.
Subsequent key normalization and RoPE are omitted for clarity.
Projection weights and caches remain depth-specific: Joint-KV shares
only across branches at the same depth, requiring synchronization
before attention.

\textbf{Cache Size.}
At fixed branch depth and KV dimensions, the Joint-KV cache in the branch region is $1/K$ that of Independent-KV, so the savings increase with the branch count. 

\textbf{Serving.}
Our inference engine, built on nano-vLLM~\citep{yu2025nanovllm}, uses the same underlying kernel backend and cache manager for
\mbt{}s and Standard.
A decode step processes the \texttt{prelude}, branch region, and
\texttt{coda} in order.
We use two additional execution mechanisms, branch grouping and head folding,
to run the branches together and reduce repeated KV reads
(Figure~\ref{fig:system-design}).

\textbf{Branch Grouping.}
Our engine groups blocks at the same depth across branches into a
batched execution stage. With Joint-KV, the engine first constructs
the shared KV cache from the branch input representations, then
computes the branch-specific attention and MoE outputs together.
Compared with executing every block sequentially, branch grouping
reduces the number of sequential block stages from
$L_{\mathrm{physical}}$ to $D=P+\max_i L_i+C$
(Eq.~\ref{eq:branch-path}).
For example, a 4-branch W3 Balanced \mbt{} contains 24 physical blocks
but requires only 12 sequential block stages with branch grouping,
compared with 16 for Standard.

\textbf{Head Folding.}
Sharing KV caches reduces storage, but separate attention calls for different branches can still reload the same KV data.
\textit{Head folding} folds the branch dimension into the query-head dimension, grouping queries by their shared KV head so that all
branches use a single grouped-query attention (GQA) call~\citep{ainslie2023gqa}.
This allows the attention kernel to reuse shared KV data across branches,
reducing redundant memory traffic.
Each query head still computes its own softmax over the same keys and
values, preserving the attention function itself.

\textbf{Distributed Execution.}
\mbt{}s can be distributed through tensor parallelism or branch parallelism.
With Joint-KV, these layouts differ in shared KV-cache construction.
With tensor parallelism, devices replicate branch hidden states
while sharding projection weights and attention heads.
Each device constructs its assigned shared KV heads locally,
without an additional collective.
Head folding also remains local when the corresponding query
heads from all branches are placed on the same device.
Alternatively, branches can run on different devices.
Each device computes its branch's contribution to the shared KV
projection, and an all-reduce combines these contributions
before attention at each depth.
The all-reduce operates on a tensor of width $2h_{\mathrm{KV}}d_h$,
avoiding an all-gather of branch representations of total width $Kd_b$.
At fixed KV dimensions, its per-token tensor size is independent
of the branch count, although communication cost depends on the
group size and topology.
This layout distributes branch weights and computation across devices
but requires synchronization at every branch depth.
This layout may be useful when branch sharding is needed to fit
the model in memory and fast interconnects are available.
\setcounter{topnumber}{2}

\section{Experiments}
\label{sec:experiments}

\subsection{Scaling Ladder Setup}
\label{sec:setup}

We pretrain standard Transformers and \mbt{}s at four scales, W1--W4, with standard models ranging from 0.80B to 2.07B total parameters (139M--394M active-body parameters).
All models use FineWeb-Edu~\citep{penedo2024fineweb}, the Llama~3 tokenizer~\citep{grattafiori2024llama3}, and a context length of 4,096 tokens.
Our scaling ladder is comparable in scale to that of Attention Residuals~\citep{kimi2026attention}, a recent industrial technical report. Our training recipes, experimental protocols, and backbone components also draw on recent technical reports~\citep{microsoft2026maithinking,dial2026pretraining,elango2026latentmoe,kimi2026k3}.
At each scale, Standard and \mbt{} models share a training-token budget of approximately 200 tokens per parameter (TPP)~\citep{microsoft2026maithinking}, using active-body parameter count and excluding embeddings and the LM head.
A brief overview of our ladder is provided in Table~\ref{tab:model-ladder}.

Our backbone uses pre-RMSNorm~\citep{zhang2019rmsnorm} blocks with full causal GQA~\citep{ainslie2023gqa}, a 2:1 query-to-KV head ratio, and a head dimension of 64. 
We apply headwise Q/K normalization, RoPE~\citep{su2021roformer}, and sigmoid gating of attention outputs~\citep{qiu2025gated}. 
Each block contains a latent MoE~\citep{elango2026latentmoe} with 64 SwiGLU experts and top-8 routing. Quantile Balancing~\citep{kimi2026k3} balances expert loads without an auxiliary balancing loss.

We train in BF16 using AdamW~\citep{loshchilov2019decoupled}, with
$\beta=(0.9,0.95)$, weight decay of 0.1, and gradient clipping at 1.0.
The learning rate warms up linearly over the first 1\% of training, then follows a cosine schedule to one tenth of its peak. Our primary peak-learning-rate
candidates are $\{1,3,5,7,10,30\}\times10^{-4}$.
We extensively tune the learning rate for the standard Transformer at each scale to establish competitive baselines.
For \mbt{}s, we tune the learning rate on a single configuration and reuse the selected value across the remaining settings.
Global batches contain 128 sequences for W1/W2 and 256 for W3/W4.
Table~\ref{tab:training-config} in Appendix~\ref{sec:training-setup}
provides the full experimental details.

\begin{table}[!ht]
  \caption{Scaling ladder and best Wide--Shallow 2-branch \mbt{}s from Figure~\ref{fig:parallel-block-ratio}.
  Standard and \mbt{} share the backbone dimensions, physical block count, and training tokens at each scale.}
  \label{tab:model-ladder}
  \centering\fontsize{9}{11}\selectfont
  \setlength{\tabcolsep}{2.5pt}
  \begin{tabular*}{\linewidth}{@{\extracolsep{\fill}}l c c c c c c c c c@{}}
    \toprule
    \multicolumn{4}{c}{Shared backbone} & Tokens & Selected \mbt{} &
    \multicolumn{2}{c}{Parameters (B / M)} & \multicolumn{2}{c}{Loss $\downarrow$} \\
    \cmidrule(lr){1-4}\cmidrule(lr){7-8}\cmidrule(l){9-10}
    Scale & $L$ & $d_{\rm model}$ & $d_{\rm ffn}$ & (B) & Allocation & Standard & \mbt{} & Standard & \mbt{} \\
    \midrule
    W1 & 12 & 1024 & 512 & 27.845 & $4/(2+2)/4$ & 0.799 / 139.2 & 0.801 / 141.3 & 2.653 & \textbf{2.622} \\
    W2 & 14 & 1152 & 576 & 41.088 & $5/(2+2)/5$ & 1.134 / 205.4 & 1.136 / 208.1 & 2.554 & \textbf{2.525} \\
    W3 & 16 & 1280 & 640 & 57.942 & $5/(3+3)/5$ & 1.555 / 289.7 & 1.558 / 293.0 & 2.471 & \textbf{2.444} \\
    W4 & 18 & 1408 & 704 & 78.841 & $6/(3+3)/6$ & 2.074 / 394.2 & 2.077 / 398.2 & 2.392 & \textbf{2.367} \\
    \bottomrule
  \end{tabular*}
  \par\vspace{3pt}
  \begin{minipage}{\linewidth}\fontsize{8}{10}\selectfont
  $L$: physical attention--MoE blocks; $P/(A+B)/C$: \texttt{prelude}/branches/\texttt{coda}.
  Parameter pairs give total (B) / active-body (M), including/excluding the tied embedding/head, respectively.
  The MoE latent width equals $d_{\rm ffn}$. Bold marks lower validation loss.
  \end{minipage}
\end{table}

\usetikzlibrary{shapes.geometric}
\begin{figure}[!t]
\centering
\begingroup
\setbox0=\hbox{\includegraphics{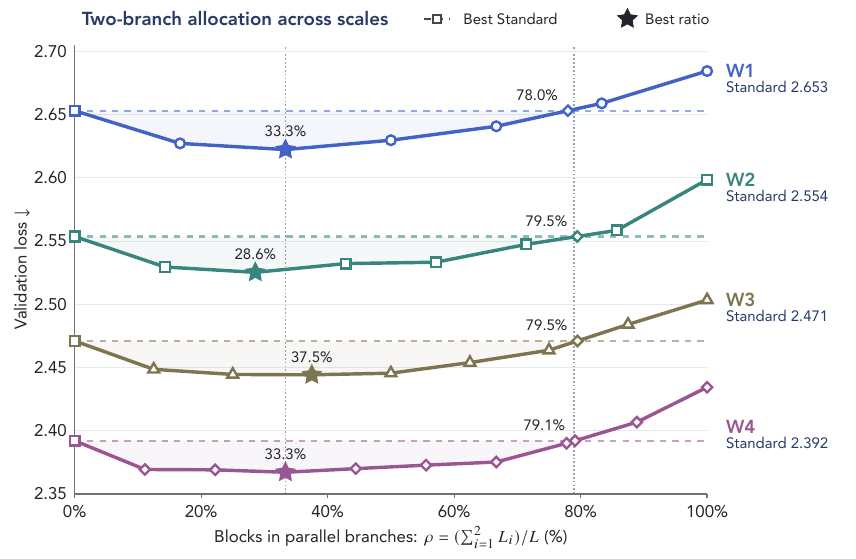}}%
\ifdim\wd0>\linewidth
  \resizebox{\linewidth}{!}{\usebox0}%
\else\usebox0\fi
\endgroup
\caption{\textbf{Scaling the parallel block ratio.} 
All four scales exhibit a consistent U-shaped trend: allocating approximately $1/3$ of blocks to parallel branches reduces loss by 0.025--0.031 relative to Standard, while increasing this fraction to approximately 79\% maintains performance on par with Standard.
}
\label{fig:parallel-block-ratio}
\end{figure}

\subsection{Scaling Transformers with Parallel Branches}
\label{sec:performance-maximizing}

We first study how allocating blocks, width, and depth across branches
affects language-model quality. In this subsection, the experiments use independent-KV caches in each branch.

\subsubsection{Scaling Laws of 2-branch MBTs}
\label{sec:parallel-block-ratio}

\textbf{Research Question.}
What fraction of a Transformer's blocks can be placed in two parallel
branches while preserving or improving model quality?

\begin{figure}[!t]
\centering
\begingroup
\includegraphics[width=0.97\linewidth]{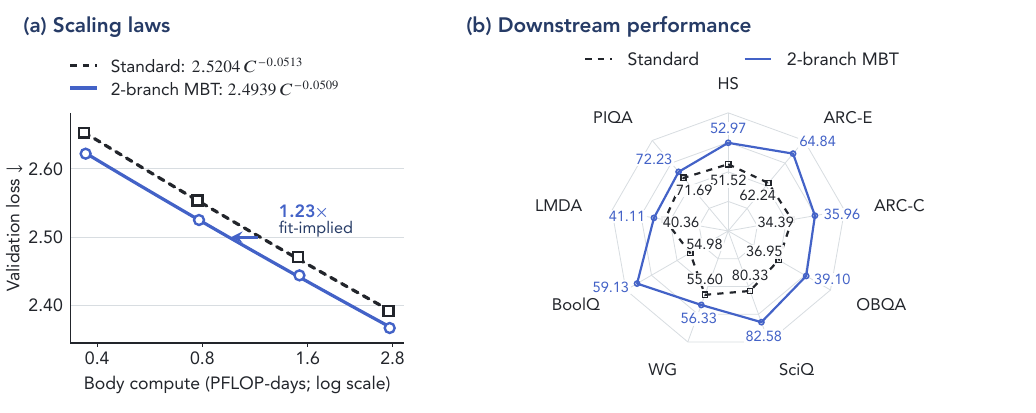}%
\endgroup
\caption{\textbf{Compute scaling and downstream performance.}
Best Wide-Shallow two-branch \mbt{}s versus Standard across W1--W4.
(a)~Validation loss versus training body compute, with power-law fits.
(b)~Zero-shot task scores averaged over W1--W4.
\mbt{}s improve compute efficiency and mean accuracy on all nine tasks.
}
\label{fig:two-branch-best}
\end{figure}

\textbf{Setup.}
Using Wide--Shallow \mbt{}s (where total training FLOPs and parameters remain almost identical to the standard reference model), we vary the fraction of physical blocks in
two equal-depth branches, $\rho=(\sum_{i=1}^2 L_i) /L$, from zero (Standard) to one
(fully branched). Within each scale, we fix the backbone width, total
physical block count, and training tokens, and split the remaining blocks
equally between the \texttt{prelude} and \texttt{coda} ($P=C$).
Branch outputs are fused by normalized concatenation followed by a linear projection.
We compare final validation loss across allocations
(Figure~\ref{fig:parallel-block-ratio}), then select the lowest-loss
allocation at each scale to study body-compute scaling and zero-shot
downstream performance (Figure~\ref{fig:two-branch-best}).
Table~\ref{tab:model-ladder} lists these selected allocations.

\textbf{Result and Interpretation.}
All four scales exhibit a highly consistent U-shaped relationship
between parallel block ratio and validation loss. The best tested
allocations cluster around $1/3$ of the blocks (28.6--37.5\%),
reducing loss by 0.025--0.031 relative to Standard.
Larger parallel regions also retain Standard-level quality: the largest
measured ratios no worse than Standard are 66.7\%, 71.4\%, 75.0\%,
and 77.8\% for W1--W4.

The sweep adds one block to each branch at a time, giving a discrete
ratio increment of $2/L$; exact equality with Standard need not fall
on a sampled allocation. We therefore linearly interpolate between
neighboring observations to estimate the equality points.
All four crossings fall within 78--80\%.
The consistent curve shape, minima near $1/3$, and tightly clustered
crossings suggest a stable block-allocation pattern across model scales.

The fitted scaling curves indicate an approximately
\textbf{1.23$\times$} body-compute advantage for the selected
two-branch \mbt{}s over Standard at matched validation loss
(Figure~\ref{fig:two-branch-best}(a)).
The downstream gains are also consistent across tasks.
Averaged over W1--W4, \mbt{} outperforms Standard on \emph{all nine}
benchmarks, raising overall mean zero-shot accuracy from 54.23\% to
56.03\% (Figure~\ref{fig:two-branch-best}(b)).

\subsubsection{Scaling Laws of multi-branch MBTs}
\label{sec:multibranch-results}

\textbf{Research Question.}
Do the language modeling gains persist beyond two branches, and how do they depend on the allocation of width and depth?

\begin{figure}[!htb]
\centering
\begingroup
\setbox0=\hbox{\includegraphics{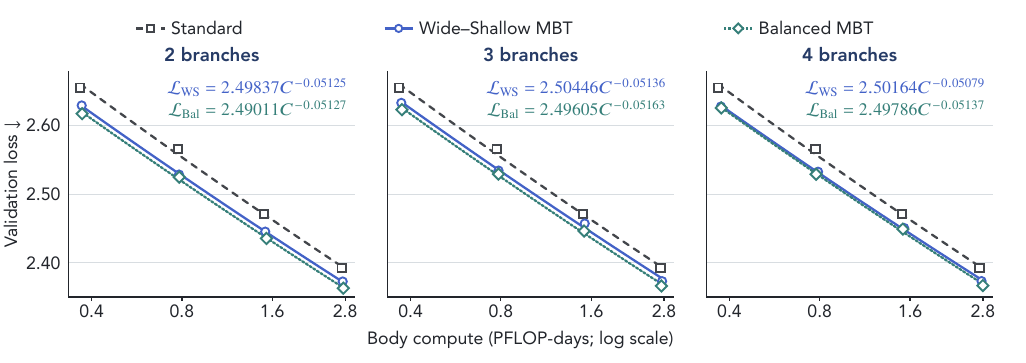}}%
\ifdim\wd0>\linewidth
  \resizebox{\linewidth}{!}{\usebox0}%
\else\usebox0\fi
\endgroup
\caption{\textbf{Scaling to more parallel branches.} Across W1--W4, increasing from 2 to 4 branches
largely preserves performance. Balanced \mbt{}s improve the performance
through joint width--depth tuning.}
\label{fig:multibranch-results}
\end{figure}

\textbf{Setup.}
We compare \mbt{}s with 2, 3, and 4 branches using the Wide--Shallow and Balanced variants defined in Figure~\ref{fig:mbt-families}, matching training compute at each scale.\footnote{Our 2-branch Narrow--Deep results
suggest benefits from retaining depth as models scale
(Section~\ref{sec:analysis-width-depth}). With 3 or 4 branches, however,
preserving full depth at fixed compute requires much smaller widths and
is accompanied by poorer quality, especially with 4 branches
(Appendix~\ref{sec:narrow-depth-count}). We therefore study Balanced MBTs,
trading some depth for greater width within the same compute budget.}
Relative to the standard model at each scale, we retain \textbf{half of the blocks} in the shared \texttt{prelude} and \texttt{coda}, keeping their block configurations unchanged, and reorganize the middle region into parallel branches.
Wide--Shallow models place half of the original blocks in
this region, preserving their configurations and distributing them
as evenly as possible across branches. Branch depths may differ
when this block count is not divisible by the number of branches.
Balanced models instead adjust branch widths and depths to match
the reference training body compute; their total physical block
count and fraction of blocks in branches can therefore differ.
Table~\ref{tab:w3-width-depth-example} illustrates an examplar W3 settings.
Figure~\ref{fig:multibranch-results} plots final loss against body compute
across W1--W4 with 7 fitted scaling curves;
Table~\ref{tab:multibranch-configs} lists all configurations.

\begin{table}[!htb]
\captionsetup{skip=4pt}
\caption{W3 width--depth allocations at comparable compute.
Full configurations appear in Table~\ref{tab:multibranch-configs}.}
\label{tab:w3-width-depth-example}
\centering\fontsize{9}{9.5}\selectfont
\renewcommand{\arraystretch}{1.00}
\setlength{\tabcolsep}{3pt}
\begin{tabular*}{\linewidth}{@{\extracolsep{\fill}}l c c r r r c r r@{}}
\toprule
Model & $K$ (\# branches) & Allocation & $d$ & $m$ & $f$ & $L/D$ & Body & Loss $\downarrow$ \\
\midrule
Standard & --- & $16$ (serial) & 1280 & 640 & 640 & 16/16 & 1.5032 & 2.471 \\
\midrule
Wide--Shallow & 2 & $4/(4+4)/4$ & 1280 & 640 & 640 & 16/12 & 1.5164 & 2.445 \\
Wide--Shallow & 3 & $4/(3+3+2)/4$ & 1280 & 640 & 640 & 16/11 & 1.5230 & 2.457 \\
Wide--Shallow & 4 & $4/(2+2+2+2)/4$ & 1280 & 640 & 640 & 16/10 & 1.5296 & 2.451 \\
\midrule
Balanced & 2 & $4/(6+6)/4$ & 1024 & 512 & 512 & 20/14 & 1.5301 & 2.436 \\
Balanced & 3 & $4/(5+5+5)/4$ & 896 & 432 & 448 & 23/13 & 1.5160 & 2.446 \\
Balanced & 4 & $4/(4+4+4+4)/4$ & 896 & 400 & 400 & 24/12 & 1.5148 & 2.449 \\
\bottomrule
\end{tabular*}
\par\vspace{2pt}
\begin{minipage}{\linewidth}\fontsize{8}{10}\selectfont
$d,m,f$: residual, MoE latent, and expert FFN widths; $L/D$: physical blocks / depth.
Body compute: PFLOP-days.
\end{minipage}
\end{table}

\textbf{Result and Interpretation.}
Increasing the branch count from 2 to 3 or 4 yields a mean absolute
change in validation loss below 0.006 across both families and all
four scales, while every configuration outperforms Standard. \mbt{}s can therefore expose more concurrent branch executions
while maintaining their quality advantage.
Balanced MBTs achieve lower loss than their Wide--Shallow counterparts
at every scale and branch count. For all 3 branch counts, the fitted law
$L=AC^{-\alpha}$ yields both a lower $A$ and a slightly larger $\alpha$
for Balanced models (Figure~\ref{fig:multibranch-results}). These results
suggest that MBTs offer a rich design space: joint width--depth tuning
can improve performance beyond fixed-width branching within a given
compute budget.

\subsection{Scaling MBTs with Joint-KV}
\label{sec:inference-speedup}

\textbf{Research Question.}
Can our final recipe, Balanced \mbt{}s with Joint-KV, simultaneously improve both
language-model quality and inference speed over standard Transformers?

\textbf{Setup.}
We reuse the Balanced \mbt{}s from
Section~\ref{sec:multibranch-results} as Independent-KV baselines,
covering 2, 3, and 4 branches across W1--W4.
For each configuration, we train a Joint-KV counterpart with the
same block allocation, widths, and training settings to isolate
the effect of KV sharing.
Table~\ref{tab:shared-kv-w4} shows the W4 setup as an example.
Full configurations and results appear in
Appendix~\ref{sec:shared-kv-details}.
We report the theoretical full-model KV-cache savings relative to
Standard at the same scale, context length, and cache precision.

\begin{table}[!htb]
\captionsetup{skip=4pt}
\caption{W4 Balanced \mbt{}s with Independent-KV and Joint-KV.
Cache capacity and savings are for Joint-KV relative to Standard;
full W1--W4 results appear in Table~\ref{tab:shared-kv-all}.}
\label{tab:shared-kv-w4}
\centering\fontsize{9}{11}\selectfont
\setlength{\tabcolsep}{2.5pt}
\renewcommand{\arraystretch}{1.05}
\begin{tabular*}{\linewidth}{@{\extracolsep{\fill}}l c r r r r r r r@{}}
\toprule
\multicolumn{2}{c}{Architecture} & \multicolumn{3}{c}{Block widths} &
\multicolumn{2}{c}{Loss $\downarrow$} &
\multicolumn{2}{c}{KV cache} \\
\cmidrule(lr){1-2}\cmidrule(lr){3-5}\cmidrule(lr){6-7}\cmidrule(l){8-9}
Model & Allocation & $d$ & $m$ & $f$ & Independent & Joint & KiB/token & Saved \\
\midrule
Standard & 18 (serial) & 1408 & 704 & 704 & 2.392 & -- & 49.50 & -- \\
$K=2$ & $4/(7+7)/4$ & 1152 & 592 & 592 & 2.363 & \textbf{2.361} & 37.75 & 23.74\% \\
$K=3$ & $4/(6+6+6)/5$ & 896 & 496 & 496 & 2.366 & 2.366 & 35.25 & 28.79\% \\
$K=4$ & $4/(6+6+6+6)/5$ & 768 & 432 & 432 & 2.367 & \textbf{2.364} & 33.75 & 31.82\% \\
\bottomrule
\end{tabular*}
\par\vspace{3pt}
\begin{minipage}{\linewidth}\fontsize{8}{10}\selectfont
$K$: number of branches. Widths apply to Standard blocks or \mbt{} branches;
shared outer blocks use Standard widths. Cache capacity uses BF16.
\end{minipage}
\end{table}

\begin{figure}[!htb]
\centering
\begingroup
\setbox0=\hbox{\includegraphics{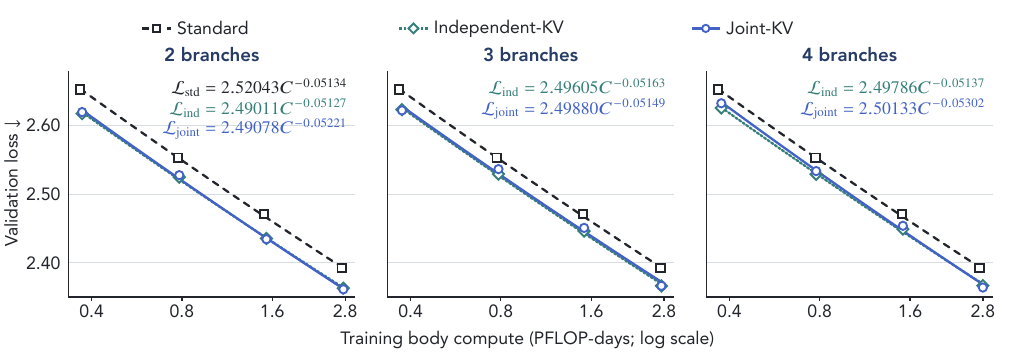}}%
\ifdim\wd0>\linewidth
  \resizebox{\linewidth}{!}{\usebox0}%
\else\usebox0\fi
\endgroup
\caption{\textbf{Joint-KV preserves strong scaling performance.} Joint-KV closely tracks
Independent-KV across branch counts and improves on Standard while reducing cache capacity.}
\label{fig:shared-kv-scaling}
\end{figure}

\textbf{Language Modeling.}
Figure~\ref{fig:shared-kv-scaling} compares final validation loss against
each variant's body compute. Joint-KV closely tracks Independent-KV
across scales and branch counts. All 12 Joint-KV models achieve lower
validation loss than Standard, retaining the gains from branching
with significantly smaller KV cache size. The gains also extend to downstream evaluation:
averaged equally over W1--W4 and the nine zero-shot tasks,
2-, 3-, and 4-branch Joint-KV \mbt{}s score 55.92\%, 55.56\%, and
55.47\%, respectively, compared with 54.23\% for Standard.
Appendix Table~\ref{tab:joint-kv-downstream} records the full quantitative results.

\textbf{Inference Efficiency.}
We benchmark decoding on one NVIDIA H200 GPU with the engine of Section~\ref{sec:system_method}, which serves Standard and \mbt{}s identically. We test the Balanced Joint-KV \mbt{}s of Table~\ref{tab:shared-kv-all} with two and four branches at all four scales. Each request has a 4K-token prompt from FineWeb-Edu~\citep{penedo2024fineweb} or a 16K-, 64K-, or 128K-token prompt from RULER~\citep{hsieh2024ruler}, and generates 1K tokens with temperature 1.0 and top-$p$ 0.95. The engine keeps $N$ requests in flight, with $N\in\{1,8,64\}$ ($\{1,8,32\}$ for 64K-token and $\{1,8,16\}$ for 128K-token prompts). We define \textit{decode speedup} as the decode throughput of an \mbt{}, \textit{i.e.}, generated tokens per second excluding prefill, divided by that of Standard. 
Each \mbt{} runs on the same GPU as its Standard baseline. We measure Standard before and after the \mbt{}s and use the mean of the two Standard runs.
On average, the two Standard measurements differ by 2.0\% at $N=1$ and 0.6\% for $N\geq8$. With the 1K generated tokens, every sequence exceeds the 4K training context, so these runs measure speed only.

\begin{figure}[!htbp]
  \centering
  \input{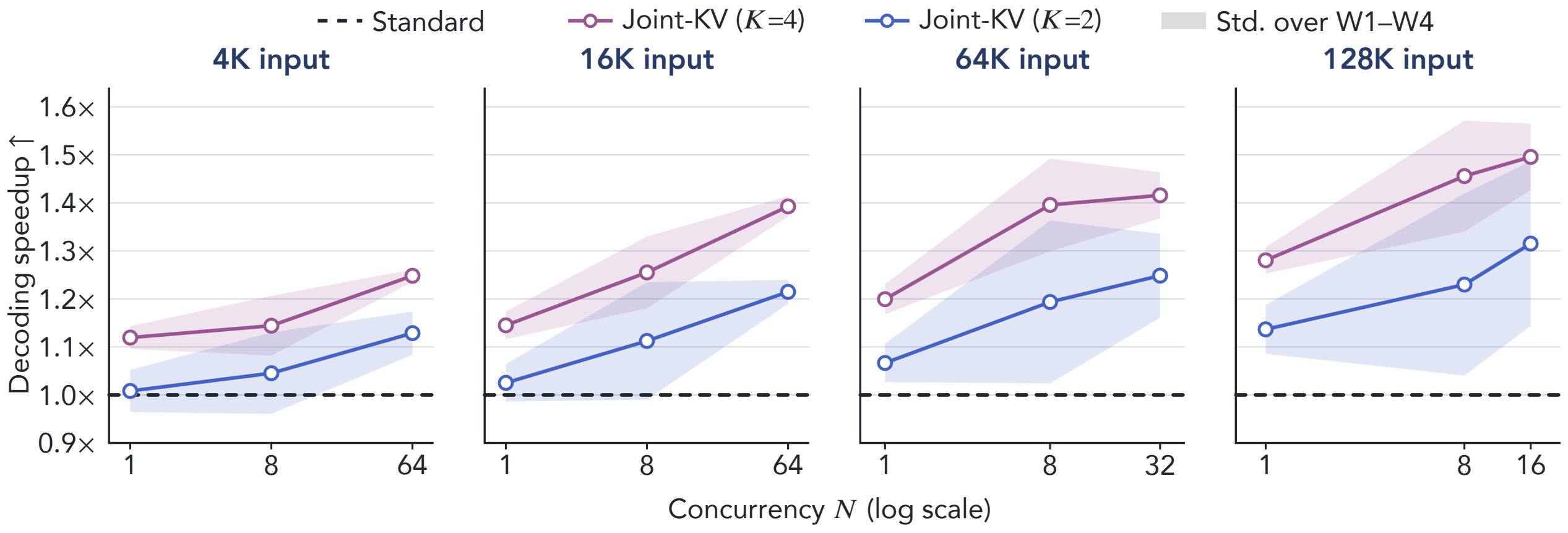}%
  \caption{\textbf{Decode speedup over Standard.} Decode throughput of Balanced Joint-KV \mbt{}s relative to Standard on one H200 GPU, averaged over scales W1--W4, as a function of the number of concurrent requests $N$, for 4K-, 16K-, 64K-, and 128K-token inputs that each generate 1K tokens. Shaded bands show one standard deviation across scales.
  }
  \label{fig:decode-speedup}
\end{figure}

Figure~\ref{fig:decode-speedup} shows the results. Averaged over the four scales, the 4-branch Joint-KV \mbt{} decodes faster than Standard in every setting. Its speedup ranges from $1.12\times$ with a single 4K-token request to $1.50\times$ with 16 concurrent 128K-token requests. It is at least $1.08\times$ faster at every scale and setting. Averaged over scales, two branches range from $1.01\times$ to $1.32\times$. Four branches outperform two at every scale and in every setting, since they need fewer serial steps and read fewer KV bytes. These speedups come without a quality cost: at every scale, both models reach lower validation loss than Standard (Table~\ref{tab:shared-kv-all}).

\textbf{Sources of the Speedup.}
The time of a decode step has two parts. The first is a fixed cost for executing the serial blocks. It depends on neither the number of requests nor the input length. Branch grouping runs the blocks of all branches at the same depth as one step, so an \mbt{} pays this cost for fewer blocks. The depth of Standard divided by that of the \mbt{} is the depth ratio $r_{\mathrm{d}}$ (Table~\ref{tab:decode-depth}). The second part is the time to load data from memory and compute on it, which grows with the concurrency and the context length. For the models we test, this part is bound by memory traffic. With Joint-KV and head folding, each depth of the branch region reads its shared KV cache once. An \mbt{} therefore reads only $1-s$ of the KV bytes of Standard, where $s$ is the KV saving of Table~\ref{tab:shared-kv-all}. We denote by $F$ the fixed cost and by $M$ the KV-read time of Standard. The speedup is then approximately
\begin{equation}
  S\approx\frac{F+M}{F/r_{\mathrm{d}}+(1-s)\,M}.
  \label{eq:decode-speedup}
\end{equation}
Our settings span 4K to 128K tokens and 1 to 64 requests. With short inputs and low concurrency, $M$ is small compared with $F$, and the speedup comes mainly from the depth ratio. As the context length grows, KV reads take most of each step, and the speedup approaches the bound $1/(1-s)$. Appendix~\ref{sec:inference-all-scales} gives the profiles and a more detailed analysis. Appendix~\ref{sec:engine-ablation} measures the contribution of each engine mechanism.

\section{Ablation Studies}
\label{sec:analysis}

\subsection{Width and Depth Ablations}
\label{sec:analysis-width-depth}

\textbf{Ablation Goal.}
We test whether the gains of branching can be explained by a shallower,
wider network, and whether they persist when Standard's depth is preserved.

\textbf{Setup.}
We retain Standard and 2-branch Wide--Shallow \mbt{} as fixed references
at each scale, with roughly half the blocks in parallel branches in the
latter. We add three controls.
\emph{Global wide--shallow} widens the entire sequential stack and matches
Wide--Shallow \mbt{}'s depth, following the conventional width--depth
trade-off~\citep{bian2025scaling}.
\emph{Local wide-core} is our localized control: it replaces the two
branches with one wider core of the same depth, preserving the \texttt{prelude}
and \texttt{coda}. Both controls test whether fewer, wider sequential blocks can recover
the gains of Wide--Shallow \mbt{} at comparable compute.
Narrow--Deep \mbt{} instead keeps the same \texttt{prelude} and \texttt{coda}, together
roughly half of Standard's blocks, and gives each of two narrower
branches the full depth of Standard's middle stack. It thus serves as a depth-matched ablation against Standard models.
Table~\ref{tab:w1-width-depth-example} gives a W1 example.
Full configurations appear in
Table~\ref{tab:width-depth-controls}.

\begin{table}[!htbp]
\centering
\caption{Example of the ablation setting at W1 with fixed data and approximately matched body compute.}
\label{tab:w1-width-depth-example}
\begingroup
\fontsize{8.5}{10}\selectfont
\setlength{\tabcolsep}{4pt}
\renewcommand{\arraystretch}{1.13}
\begin{tabular}{@{}lccccc@{}}
\toprule
Family & Allocation & \shortstack{Blocks /\\depth}
  & \shortstack{Outer or all-layer\\$d/m/f$}
  & \shortstack{Core or branch\\$d/m/f$}
  & \shortstack{Body compute\\(PFLOP-days)} \\
\midrule
Standard & $12$ & $12/12$ & $1024/512/512$ & --- & 0.3665 \\
Global wide--shallow & $9$ & $9/9$ & $1280/640/512$ & --- & 0.3721 \\
Local wide-core & $3/3/3$ & $9/9$ & $1024/512/512$ & $1664/640/640$ & 0.3636 \\
Wide--Shallow \mbt{} & $3/(3+3)/3$ & $12/9$ & $1024/512/512$ & $1024/512/512$ & 0.3706 \\
Narrow--Deep \mbt{} & $3/(6+6)/3$ & $18/12$ & $1024/512/512$ & $640/320/416$ & 0.3705 \\
\bottomrule
\end{tabular}
\endgroup
\end{table}

\begin{figure}[tbp]
\centering
\begingroup
\setbox0=\hbox{\includegraphics{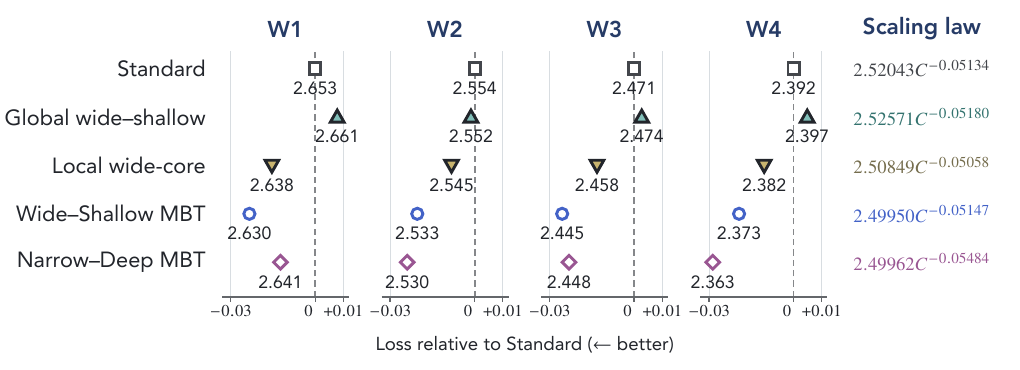}}%
\ifdim\wd0>\linewidth
  \resizebox{\linewidth}{!}{\usebox0}%
\else\usebox0\fi
\endgroup
\caption{\textbf{Ablation studies on branching.}
Positions show loss differences from Standard; labels give absolute loss.
The right column fits the scaling laws. We find branching is the key to performance improvement.}
\label{fig:width-depth-ablation}
\end{figure}

\textbf{Result and Interpretation.}
Figure~\ref{fig:width-depth-ablation} shows that Wide--Shallow \mbt{} outperforms both widening baselines (Global Wide--Shallow and Local Wide--Core) at all four scales, with a consistently better-fitted scaling curve.
Narrow--Deep \mbt{} also outperforms Standard across all scales and exhibits the steepest scaling exponent ($\alpha=0.0548$, versus $0.0515$ for
Wide--Shallow), suggesting that branching improves modeling quality even when the original sequential depth is preserved.
Together, these results demonstrate benefits beyond sequential widening at matched depth, while indicating that the 
best \mbt{} can be tuned by jointly considering width and depth.

\subsection{KV Sharing Across Branches}
\label{sec:analysis-joint-gqa}

\textbf{Ablation Goal.}
At the same KV-cache size and training budget, does constructing shared keys and values
jointly across branches (\textbf{Joint-KV}) improve modeling quality over keeping smaller,
independent caches in each branch?

\begin{table}[!b]
\centering
\captionsetup{skip=4pt}
\caption{Joint-KV versus Independent-KV across W1--W4 at matched KV-cache capacity. Each pair shares block allocations, widths, query heads, and training tokens. Training body compute is in PFLOP-days; full configurations in Table~\ref{tab:joint-gqa-ladder}.}
\label{tab:joint-gqa-setup}
\begingroup
\fontsize{8.5}{10.5}\selectfont
\setlength{\tabcolsep}{2.5pt}
\renewcommand{\arraystretch}{1.16}
\begin{tabular*}{\linewidth}{@{\extracolsep{\fill}}lcccccrrrrr@{}}
\toprule
 & & \multicolumn{3}{c}{Q/KV heads} & & \multicolumn{2}{c}{Body compute} & \multicolumn{3}{c}{Validation loss $\downarrow$} \\
\cmidrule(lr){3-5}\cmidrule(lr){7-8}\cmidrule(lr){9-11}
Scale & Allocation & Outer (ratio) & Independent-KV & Joint-KV
 & \shortstack{KV/token\\(KiB)} & Independent-KV & Joint-KV & Independent-KV & Joint-KV & $\Delta$ \\
\midrule
W1 & $3/(5+5)/3$ & $16/2\;(8{:}1)$ & $16/1$ & $16/2$ & 5.5 & 0.3848 & 0.3867 & 2.657 & \textbf{2.646} & $-0.011$ \\
W2 & $3/(6+6)/3$ & $18/2\;(9{:}1)$ & $16/1$ & $16/2$ & 6.0 & 0.7723 & 0.7763 & 2.568 & \textbf{2.557} & $-0.011$ \\
W3 & $4/(6+6)/4$ & $20/2\;(10{:}1)$ & $16/1$ & $16/2$ & 7.0 & 1.4436 & 1.4499 & 2.499 & \textbf{2.487} & $-0.012$ \\
W4 & $4/(7+7)/4$ & $24/3\;(8{:}1)$ & $18/1$ & $18/2$ & 9.5 & 2.6551 & 2.6664 & 2.392 & \textbf{2.380} & $-0.012$ \\
\bottomrule
\end{tabular*}
\par\vspace{3pt}
\begin{minipage}{\linewidth}
\fontsize{8}{9.5}\selectfont
Joint-KV's KV heads are shared by both branches; Independent-KV heads are per branch, so both store 2 KV heads per branch depth.
Independent-KV here keeps 1 KV head per branch to match Joint-KV's cache, unlike the Independent-KV models in Table~\ref{tab:shared-kv-w4}, which use the main $2{:}1$ recipe.
\end{minipage}
\endgroup
\end{table}

\textbf{Setup.}
We start from the 2-branch Balanced \mbt{}s at W1--W4.
On the main ladder, the shared KV heads cannot always be split evenly
between two branches (e.g., W2 and W4), so an exactly cache-matched
Independent-KV control does not exist at every scale.
We therefore keep block allocations and widths and re-derive the head
counts so that every pair can be matched exactly.
In doing so, we also raise the GQA ratio from the $2{:}1$ of the main
recipe to $8{:}1$ or higher (Table~\ref{tab:joint-gqa-setup}), since how KV heads are allocated across
branches matters most when few are available.
We compare two constructions with \textbf{identical KV-cache size}
(2 KV heads per branch depth in total):
\textbf{Independent-KV} gives each branch 1 private KV head;
\textbf{Joint-KV} constructs 2 KV heads per branch depth, shared by both
branches.
All other settings, including training tokens and LR candidates, are
identical within each pair (Appendix~\ref{sec:joint-gqa-details}).\looseness=-1

\textbf{Result and Interpretation.}
Joint-KV achieves lower validation loss at all four scales, by
0.011--0.012 (Table~\ref{tab:joint-gqa-setup}). Its larger KV projection
adds less than 0.6\% body compute, far too little to explain the gap.
This result shows that, at a fixed cache size, sharing KV across
branches is more effective than keeping a private KV cache in each branch.

\subsection{Fusion Mechanisms}
\label{sec:analysis-fusion}

\textbf{Ablation Goal and Setup.}
We test whether fusion needs a learned projection and whether more expressive operators improve \mbt{}. We fix the setting to be Wide-Shallow 2-branch \mbt{} of the configuration $4/(4+4)/4$ at scale W3 (half of the parameters in parallel branches), changing only the fusion operator.
Since fusing multiple vectors is extensively used in multimodal research, 
we adapt gating, joint-representation, and
bilinear mechanisms from highly-cited existing papers
\citep{arevalo2017gated,ngiam2011multimodal,kim2017hadamard}, and also test
joint SwiGLU and tokenwise softmax aggregation.
We additionally test a normalized sum, $(h_1+h_2)/\sqrt{2}$,
whose aggregation introduces no learned parameters beyond the existing
branch RMSNorms.
Table~\ref{tab:fusion-operators} gives the formulations and results.

\begin{table}[hbp]
\centering
\caption{Fusion of normalized branch vectors $h_1,h_2$ in W3 $4/(4+4)/4$.
$W=[W_1\;W_2]$, $A_i$, $B_i$, and $P$ are learned weight matrices;
$a_i=A_iW_i h_i$, $b_i=B_iW_i h_i$, and $\phi=\operatorname{SiLU}$.
$g$ is a channelwise sigmoid gate and $\beta$ is a tokenwise softmax over branches;
$\odot$ denotes elementwise multiplication.
The last row tests parameter-free aggregation.}
\label{tab:fusion-operators}
\begingroup
\fontsize{9}{11}\selectfont
\setlength{\tabcolsep}{4pt}
\renewcommand{\arraystretch}{1.25}
\begin{tabularx}{\linewidth}{@{}l>{\raggedright\arraybackslash}Xr@{}}
\toprule
Fusion & Fused output $z$ & Loss $\downarrow$ \\
\midrule
Concat--projection~\citep{feichtenhofer2016fusion}
& $W[h_1;h_2]$ & 2.445 \\
Channel competition~\citep{arevalo2017gated}
& $2[g\odot W_1h_1+(1-g)\odot W_2h_2]$ & 2.446 \\
Joint MLP~\citep{ngiam2011multimodal}
& $W[h_1;h_2]+P\phi(a_1+a_2)$ & 2.445 \\
Low-rank bilinear~\citep{kim2017hadamard}
& $W[h_1;h_2]+P(a_1\odot a_2)$ & 2.446 \\
Joint SwiGLU~\citep{shazeer2020glu}
& $W[h_1;h_2]+P[\phi(a_1+a_2)\odot(b_1+b_2)]$ & 2.444 \\
Token softmax~\citep{chen2025parallel}
& $2(\beta_1W_1h_1+\beta_2W_2h_2)$ & 2.448 \\
Normalized sum (no projection)
& $(h_1+h_2)/\sqrt{2}$ & 2.458 \\
\bottomrule
\end{tabularx}
\endgroup
\end{table}

\textbf{Result and Interpretation.}
Across the six learned fusion operators, the final loss varies by less than 0.005.
The projection-free normalized sum reaches 2.458, compared with 2.445
for concat--projection. We therefore retain concat--projection as a
simple, effective default: removing the projection hurts quality in
this setting, while more elaborate learned fusion offers little benefit.

\subsection{Branch Placement}
\label{sec:analysis-placement}

\textbf{Ablation Goal and Setup.}
We test \emph{where to branch} by placing the parallel blocks in the early, middle (ours), or late layers. We use a 2-branch Wide--Shallow \mbt{} at scale W3, with 16 blocks in total, 8 of which are distributed across the parallel branches.

\begin{figure}[htbp]
\centering
\begingroup
\setbox0=\hbox{\includegraphics{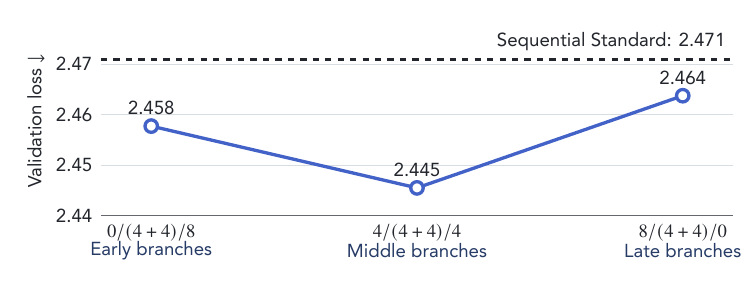}}%
\ifdim\wd0>\linewidth
  \resizebox{\linewidth}{!}{\usebox0}%
\else\usebox0\fi
\endgroup
\caption{\textbf{Branching in the middle works best.}
At W3, we place 8 shared blocks around two 4-block branches while holding
all other settings fixed. Shared processing before and after the branches
gives the lowest loss. 
}
\label{fig:branch-placement}
\end{figure}

\textbf{Result and Interpretation.}
As shown in Figure~\ref{fig:branch-placement}, middle placement achieves the lowest loss (2.445), outperforming both early (2.458) and late (2.464) branching.
The advantage of early over late branching suggests that retaining a shared \texttt{coda} is more beneficial than retaining a shared \texttt{prelude} in this setting.
Overall, these results favor placing the parallel region in the middle layers.

\section{Related Work}
\label{sec:related}

\textbf{Parallelism in large language models.}
Architectural work on language models explores concurrency within individual blocks, as in PaLM~\citep{chowdhery2022palm}, width--depth trade-offs in sequential stacks~\citep{tay2022scale,mcleish2025gemstones,bian2025scaling}, and parallel passes through a shared backbone with learned prefixes, as in ParScale~\citep{chen2025parallel}.
From a systems perspective, parallelism supports efficient execution across devices: tensor and pipeline parallelism distribute model computation~\citep{narayanan2021efficient}, and track parallelism uses narrower stacks with periodic synchronization to reduce communication overhead~\citep{wang2026paralleltrack}.
Our approach explores an architectural design axis orthogonal to those studied in prior work on parallel language models, while remaining compatible with existing systems approaches.
More broadly, multi-branch architectures have a long history in computer vision, with prominent examples including Inception~\citep{szegedy2015going}, ResNeXt~\citep{xie2017resnext}, FractalNet~\citep{larsson2017fractalnet}, and HRNet~\citep{sun2019deep}.
Earlier NLP work (Pre-LLM era) explored parallel branches within attention modules~\citep{ahmed2017weighted,fan2020multibranch} and encoder backbones~\citep{burtsev2021multi}.
However, parallel composition of full Transformer blocks remains underexplored as a means of scaling modern autoregressive language models.

\textbf{Information flow and KV-cache sharing.}
Recent work improves information flow through expanded residual streams and selective aggregation of representations across depth~\citep{zhu2024hyperconnections,xie2025mhc,kimi2026attention,zeng2026depthattention,zhu2026moda}.
These approaches enrich interactions within a sequential block stack, although maintaining and mixing additional representations can introduce extra memory traffic.
\mbt{} takes a different approach by reorganizing full Transformer blocks into parallel branches and fusing their outputs before subsequent shared layers.
Prior work shares keys and values across attention heads~\citep{shazeer2019fast,ainslie2023gqa} or across decoder layers, as in Cross-Layer Attention and YOCO~\citep{brandon2024crosslayer,sun2024yoco}.
Joint-KV shares along a different axis that exists only in multi-branch models: across parallel branches at the same depth, while retaining branch-specific queries.
This axis is orthogonal to head and layer sharing (our models already combine it with GQA) and enables head folding to reduce redundant KV reads during long-context decoding.
Our results show that the resulting cache savings and inference gains coexist with stronger language modeling performance than standard Transformers.

\section{Conclusion}
\label{sec:conclusion}
We introduced \mbt{}s, which organize full Transformer blocks into
parallel branches between shared initial and final layers.
Across four MoE model scales, branching improves language-model
quality and fitted scaling at comparable training compute.
Balanced branch allocations, Joint-KV, and an inference engine
that exploits branch structure together enable better modeling
quality and faster decoding. These results establish branching
as a practical architectural choice alongside width and depth.
These findings open several directions for future work:
\begin{itemize}[beginpenalty=10000]
    \clubpenalty=10000
    \item \textbf{Joint Scaling Laws with Branching.}
    Existing scaling studies guide model and training configurations
    under a given compute budget~\citep{hoffmann2022training,held2026delphi}.
    Extending this framework to jointly account for branch count,
    branch depth, and hidden dimensions could identify how best to
    allocate parameters within that budget. Incorporating inference
    cost would further guide configurations that balance modeling
    quality and serving efficiency.

    \item \textbf{\mbt{} + YOCO + Cross-Layer Attention.}
    YOCO enables aggressive cache reduction and early-exit
    prefill~\citep{sun2024yoco}, while Cross-Layer Attention shares KV
    across neighboring layers~\citep{brandon2024crosslayer}.
    Recent work such as HySparse2 combines selective sharing across
    layers with sparse attention~\citep{wei2026hysparse2}.
    Combining these approaches with \mbt{} would enable joint design
    of KV sharing across branches and layers, potentially improving
    the trade-off between modeling quality, prefill efficiency,
    and decoding speed.

    \item \textbf{MBTs for Multimodality.}
    Multimodal Transformers such as MM-DiT already use distinct parameters
    for different modalities~\citep{esser2024scaling}.
    MBTs could extend this idea to branches that serve as text, image,
    or audio backbones, with widths, depths, and block types tailored
    to each modality. Studying where these branches should interact
    and fuse could improve the balance between modality specialization
    and cross-modal information exchange.
\end{itemize}

\clearpage
\bibliographystyle{unsrtnat}
\bibliography{references/references}

\clearpage
\appendix
\section{Additional Experimental Details}
\label{sec:experiment-details}

We first give general training and evaluation settings, then expand
the corresponding main-text experiments and analyses in the same order.

\subsection{Architecture and pretraining setup}
\label{sec:training-setup}

\textbf{General settings.}
This subsection expands the common training recipe in
Section~\ref{sec:setup} and the ladder in Table~\ref{tab:model-ladder}.
Table~\ref{tab:training-config} lists the Standard backbone at each scale.
Wide--Shallow \mbt{}s retain these block dimensions; Narrow--Deep and Balanced
\mbt{}s change the branch widths and depths, as listed with their results.
All variants retain the training-token budget of their Standard reference.
Here $m$ is the shared MoE latent width and $f$ is each expert's SwiGLU
hidden width; neither is the residual width $d$.

\begin{table}[!htbp]
  \caption{Standard backbone and pretraining configuration for W1--W4.
  \mbt{} variants use the same per-scale data and optimization recipe unless
  their experiment specifies a change.}
  \label{tab:training-config}
  \centering\fontsize{9}{11}\selectfont
  \renewcommand{\arraystretch}{1.10}
  \setlength{\tabcolsep}{5pt}
  \begin{tabularx}{\linewidth}{@{}>{\raggedright\arraybackslash}p{0.37\linewidth} Z Z Z Z@{}}
    \toprule
    Setting & W1 & W2 & W3 & W4 \\
    \midrule
    \multicolumn{5}{@{}l}{\textbf{Backbone}} \\
    Blocks $L$ & 12 & 14 & 16 & 18 \\
    Residual width $d$ & 1,024 & 1,152 & 1,280 & 1,408 \\
    MoE latent width $m$ & 512 & 576 & 640 & 704 \\
    Expert hidden width $f$ & 512 & 576 & 640 & 704 \\
    Query / KV heads & 16 / 8 & 18 / 9 & 20 / 10 & 22 / 11 \\
    Attention head dimension & 64 & 64 & 64 & 64 \\
    Experts / active experts per token & 64 / 8 & 64 / 8 & 64 / 8 & 64 / 8 \\
    Total parameters (B) & 0.7990 & 1.1335 & 1.5549 & 2.0735 \\
    Active-body parameters (M) & 139.226 & 205.441 & 289.713 & 394.204 \\
    \midrule
    \multicolumn{5}{@{}l}{\textbf{Training schedule}} \\
    Training tokens (B) & 27.845 & 41.088 & 57.942 & 78.841 \\
    Context length & 4,096 & 4,096 & 4,096 & 4,096 \\
    Global batch (sequences) & 128 & 128 & 256 & 256 \\
    Optimizer updates $T$ & 53,110 & 78,370 & 55,258 & 75,189 \\
    Warmup updates & 531 & 783 & 552 & 751 \\
    \midrule
    \multicolumn{5}{@{}l}{\textbf{Shared settings}} \\
    Optimizer & \multicolumn{4}{c}{AdamW; $\beta=(0.9,0.95)$, $\epsilon=10^{-8}$} \\
    Weight decay / gradient clipping & \multicolumn{4}{c}{0.1 / 1.0} \\
    Learning-rate schedule & \multicolumn{4}{c}{Linear warmup; cosine decay to $0.1\eta_{\max}$} \\
    Normalization & \multicolumn{4}{c}{Pre-RMSNorm and Q/K RMSNorm; $\epsilon=10^{-5}$} \\
    Position embedding & \multicolumn{4}{c}{Full-channel RoPE; base 500,000} \\
    Vocabulary / embedding & \multicolumn{4}{c}{128,256; tied input embedding and LM head} \\
    Precision / seed & \multicolumn{4}{c}{BF16 / 1234} \\
    Attention / hidden dropout & \multicolumn{4}{c}{0 / 0} \\
    \bottomrule
  \end{tabularx}
  \par\vspace{3pt}
  \begin{minipage}{\linewidth}\fontsize{8}{10}\selectfont
  Total parameters include the tied embedding/head once; active-body
  parameters exclude it and count the selected experts.
  Token budgets are approximately $200\times$ Standard active-body parameters,
  rounded to a complete optimizer update. Widths and depths of modified
  branches are specified in their experiment tables.
  \end{minipage}
\end{table}

\textbf{Attention and experts.}
Attention uses learned headwise RMSNorm on queries and keys before RoPE,
and a channelwise sigmoid gate on the attention output before its output
projection~\citep{qiu2025gated}. The MoE router reads the normalized
$d$-dimensional residual state. Shared projections map $d\to m$ before
expert computation and $m\to d$ after expert combination. Each expert has
two $m\to f$ SwiGLU input maps and one $f\to m$ output map.
Quantile Balancing~\citep{kimi2026k3} updates expert-selection biases once
per optimizer step using statistics accumulated over the global batch
with 10,000 histogram bins. Combination weights use the original,
unbiased sigmoid scores of the selected experts, renormalized to sum to
2.5. Routing arithmetic is FP32; no tokens are dropped, and there are no
shared experts, auxiliary balancing loss, or router z-loss.

\textbf{Data and tokenization.}
We use the same pretokenized FineWeb-Edu release across comparisons, with
64 equal-weight training shards in a fixed order and the Llama~3 vocabulary
of 128,256 tokens. Documents end with an end-of-document token and are
packed into 4,096-token training sequences with next-token labels.
Attention and positions are not reset at document boundaries, and
end-of-document tokens contribute to the loss. Within each scale,
comparisons retain the same data prefix and sample-order construction.
Validation uses a separate, fixed prefix of 12,160 sequences
(49,807,360 prediction positions), excluded from the training split.

\textbf{Optimization and initialization.}
Training uses Megatron-LM~\citep{narayanan2021efficient} with BF16 weights
and FP32 optimizer state. AdamW excludes normalization scales and biases
from weight decay but applies it to the tied embedding matrix. Base
weights use a normal initialization with standard deviation 0.02;
normalization gains start at one. In the reference backbone, attention
output, expert output, and shared latent up-projection weights use
$0.02/\sqrt{2L}$. The input embedding and LM head share weights.
We keep the global batch fixed when changing the microbatch size or
number of gradient-accumulation steps.

\textbf{Learning-rate selection.}
Our primary peak-learning-rate candidates are
$\{1,3,5,7,10,30\}\times10^{-4}$, as in Section~\ref{sec:setup}.
We select checkpoints by final validation loss. The selected Standard
reference uses $5\times10^{-4}$ at all four scales (larger-scaled model has larger batch sizes).
Architecture sweeps use $7\times10^{-4}$ for W1/W2 and
$5\times10^{-4}$ for W3/W4.
Warmup starts at zero and lasts $\lfloor0.01T\rfloor$ optimizer updates,
followed by cosine decay to $0.1\eta_{\max}$ at the full horizon $T$.
All runs use seed 1234.
\subsection{Evaluation and scaling fits}
\label{sec:evaluation-protocol}

This subsection details the downstream evaluation in
Sections~\ref{sec:parallel-block-ratio} and~\ref{sec:inference-speedup}
(Figure~\ref{fig:two-branch-best} and Table~\ref{tab:joint-kv-downstream})
and the compute-accounting and fitting conventions used throughout
Sections~\ref{sec:experiments} and~\ref{sec:analysis}.

\textbf{Downstream evaluation.}
We use lm-evaluation-harness v0.4.12 with frozen task
definitions and dataset revisions. All nine tasks are zero-shot and
use the native BF16 model with a 4,096-token context limit. Requests of
equal length are batched without padding; batch shapes that fail
numerical equivalence checks use single-request evaluation. When a
request exceeds the context limit, only the left context is truncated,
preserving the scored continuation.
The nine tasks are HellaSwag (HS); the Easy and Challenge subsets of the
AI2 Reasoning Challenge (ARC-E and ARC-C); OpenBookQA (OBQA); SciQ;
WinoGrande (WG); Boolean Questions (BoolQ); LAMBADA (LMDA); and
Physical Interaction: Question Answering (PIQA).
We report length-normalized accuracy (\texttt{acc\_norm}) for HS,
ARC-E/C, OBQA, SciQ, and PIQA, and accuracy (\texttt{acc}) for WG,
BoolQ, and LMDA. LMDA uses the \texttt{lambada\_openai} task.
Tables~\ref{tab:results} and~\ref{tab:joint-kv-downstream} list the per-scale scores. Figure~\ref{fig:two-branch-best}(b)
averages each task equally over W1--W4; Average weights the nine tasks
equally. All averages are calculated before rounding.

\textbf{Compute and scaling fits.}
Body compute excludes the tied LM head and includes attention, MoE,
width bridges, and fusion projections. We count two FLOPs per
multiply--add and approximate training matrix work by three times the
forward work, including causal attention at sequence length 4,096.
Elementwise operations, activation recomputation, optimizer updates,
communication, and evaluation are excluded.
One PFLOP-day equals $8.64\times10^{19}$ FLOPs.
For each model family, we fit $\mathcal{L}(C)=AC^{-\alpha}$ by unweighted
least squares in log compute and log loss using the four final
W1--W4 checkpoints. Curves are drawn over their measured compute ranges.
The $1.23\times$ comparison in Figure~\ref{fig:two-branch-best}(a) is
obtained by inverting the two fits at loss 2.500, within their common
measured range.

\textbf{Radar axes.}
Each spoke in Figure~\ref{fig:two-branch-best}(b) spans 8 percentage
points, with rings spaced 2 points apart. Clockwise from HS, the lower
bounds are 47, 58, 30, 33, 76, 51, 52, 36, and 67 for HS, ARC-E,
ARC-C, OBQA, SciQ, WG, BoolQ, LMDA, and PIQA, respectively.
Point labels show the original percentage scores.

\subsection{Two-branch downstream results}
\label{sec:individual-downstream}
For Section~\ref{sec:parallel-block-ratio}, Table~\ref{tab:results}
reports the per-scale results underlying
Figure~\ref{fig:two-branch-best}(b), using the selected allocations in
Table~\ref{tab:model-ladder}.
\begin{table}[!htbp]
\centering
\caption{Per-scale zero-shot results for the Standard and selected 2-branch
\mbt{} checkpoints in Figure~\ref{fig:two-branch-best}.
Scores are percentages ($\uparrow$); the higher score in each pair is bold.}
\label{tab:results}
\begingroup
\fontsize{9}{11}\selectfont
\setlength{\tabcolsep}{4pt}
\renewcommand{\arraystretch}{1.12}
\begin{tabular*}{\linewidth}{@{\extracolsep{\fill}}l*{8}{r}@{}}
\toprule
& \multicolumn{2}{c}{W1} & \multicolumn{2}{c}{W2} & \multicolumn{2}{c}{W3} & \multicolumn{2}{c}{W4} \\
\cmidrule(lr){2-3}\cmidrule(lr){4-5}\cmidrule(lr){6-7}\cmidrule(lr){8-9}
Task & Standard & \mbt{} & Standard & \mbt{} & Standard & \mbt{} & Standard & \mbt{} \\
\midrule
HellaSwag & 45.15 & \textbf{46.44} & 50.29 & \textbf{52.15} & 53.30 & \textbf{54.65} & 57.34 & \textbf{58.63} \\
ARC-E & 57.37 & \textbf{60.82} & 62.46 & \textbf{63.72} & 63.47 & \textbf{67.17} & 65.66 & \textbf{67.63} \\
ARC-C & 31.06 & \textbf{31.40} & 33.45 & \textbf{35.15} & 35.58 & \textbf{38.91} & 37.46 & \textbf{38.40} \\
OBQA & 34.40 & \textbf{36.20} & 35.20 & \textbf{39.20} & 37.60 & \textbf{39.20} & 40.60 & \textbf{41.80} \\
SciQ & 75.20 & \textbf{77.00} & 80.50 & \textbf{82.50} & 84.50 & \textbf{84.70} & 81.10 & \textbf{86.10} \\
WG & \textbf{54.85} & 54.38 & 53.28 & \textbf{55.33} & \textbf{57.85} & 57.62 & 56.43 & \textbf{58.01} \\
BoolQ & 56.36 & \textbf{58.65} & 47.22 & \textbf{58.07} & 57.55 & \textbf{58.84} & 58.78 & \textbf{60.95} \\
LMDA & \textbf{37.42} & 37.05 & 38.95 & \textbf{39.47} & 40.56 & \textbf{41.84} & 44.52 & \textbf{46.07} \\
PIQA & \textbf{69.80} & 69.53 & 71.00 & \textbf{71.65} & \textbf{72.36} & 72.25 & 73.61 & \textbf{75.46} \\
\midrule
Average & 51.29 & \textbf{52.39} & 52.48 & \textbf{55.25} & 55.86 & \textbf{57.24} & 57.28 & \textbf{59.23} \\
\bottomrule
\end{tabular*}
\endgroup
\par\vspace{3pt}
\begin{minipage}{\linewidth}\fontsize{8}{10}\selectfont
We report normalized accuracy for HellaSwag, ARC-E/C, OBQA, SciQ, and PIQA;
accuracy for WG, BoolQ, and LMDA (LAMBADA).
Average weights the nine tasks equally. All metrics use lm-eval 0.4.12.
\end{minipage}
\end{table}

\subsection{Multi-branch configurations}
\label{sec:multibranch-details}
For Section~\ref{sec:multibranch-results} and Figure~\ref{fig:multibranch-results},
Table~\ref{tab:multibranch-configs} expands the W3 example in
Table~\ref{tab:w3-width-depth-example} to all four scales.
Shared-layer dimensions and training tokens follow
Table~\ref{tab:training-config}.
\begin{table}[!htb]
\captionsetup{skip=4pt}
\caption{Branch-count configurations for Figure~\ref{fig:multibranch-results}.
$L/D$ denotes physical block count / depth; $d/m/f$ gives
branch residual / MoE latent / expert FFN widths.
Compute is body / total PFLOP-days.}
\label{tab:multibranch-configs}
\centering\fontsize{8}{10}\selectfont
\setlength{\tabcolsep}{3.0pt}
\renewcommand{\arraystretch}{1.00}
\begin{tabular*}{\linewidth}{@{\extracolsep{\fill}}llc c c c c r@{}}
\toprule
Scale & \mbt{} family & $K$ & $P/(\text{branches})/C$ & $d/m/f$ & $L/D$ & Body / Total & Loss $\downarrow$\\
\midrule
W1 & Wide--Shallow & 2 & $3/(3+3)/3$ & 1024/512/512 & 12/9 & 0.3706 / 0.6245 & 2.630 \\
W1 & Wide--Shallow & 3 & $3/(2+2+2)/3$ & 1024/512/512 & 12/8 & 0.3726 / 0.6266 & 2.633 \\
W1 & Wide--Shallow & 4 & $3/(2+2+1+1)/3$ & 1024/512/512 & 12/8 & 0.3746 / 0.6286 & 2.628 \\
W1 & Balanced & 2 & $3/(5+5)/3$ & 768/368/384 & 16/11 & 0.3717 / 0.6257 & 2.618 \\
W1 & Balanced & 3 & $3/(4+4+4)/3$ & 640/368/368 & 18/10 & 0.3745 / 0.6284 & 2.623 \\
W1 & Balanced & 4 & $3/(3+3+3+3)/3$ & 640/368/368 & 18/9 & 0.3758 / 0.6297 & 2.626 \\
\midrule
W2 & Wide--Shallow & 2 & $3/(4+3)/4$ & 1152/576/576 & 14/11 & 0.7822 / 1.2038 & 2.529 \\
W2 & Wide--Shallow & 3 & $3/(2+3+2)/4$ & 1152/576/576 & 14/10 & 0.7860 / 1.2076 & 2.535 \\
W2 & Wide--Shallow & 4 & $3/(2+2+2+1)/4$ & 1152/576/576 & 14/9 & 0.7898 / 1.2114 & 2.533 \\
W2 & Balanced & 2 & $3/(6+6)/3$ & 896/464/464 & 18/12 & 0.7838 / 1.2054 & 2.525 \\
W2 & Balanced & 3 & $3/(5+5+5)/4$ & 768/352/352 & 22/12 & 0.7856 / 1.2072 & 2.529 \\
W2 & Balanced & 4 & $3/(4+4+4+4)/4$ & 640/384/384 & 23/11 & 0.7782 / 1.1998 & 2.529 \\
\midrule
W3 & Wide--Shallow & 2 & $4/(4+4)/4$ & 1280/640/640 & 16/12 & 1.5164 / 2.1769 & 2.445 \\
W3 & Wide--Shallow & 3 & $4/(3+3+2)/4$ & 1280/640/640 & 16/11 & 1.5230 / 2.1835 & 2.457 \\
W3 & Wide--Shallow & 4 & $4/(2+2+2+2)/4$ & 1280/640/640 & 16/10 & 1.5296 / 2.1901 & 2.451 \\
W3 & Balanced & 2 & $4/(6+6)/4$ & 1024/512/512 & 20/14 & 1.5301 / 2.1907 & 2.436 \\
W3 & Balanced & 3 & $4/(5+5+5)/4$ & 896/432/448 & 23/13 & 1.5160 / 2.1765 & 2.446 \\
W3 & Balanced & 4 & $4/(4+4+4+4)/4$ & 896/400/400 & 24/12 & 1.5148 / 2.1754 & 2.449 \\
\midrule
W4 & Wide--Shallow & 2 & $4/(5+4)/5$ & 1408/704/704 & 18/14 & 2.7482 / 3.7369 & 2.373 \\
W4 & Wide--Shallow & 3 & $4/(3+3+3)/5$ & 1408/704/704 & 18/12 & 2.7591 / 3.7478 & 2.373 \\
W4 & Wide--Shallow & 4 & $4/(3+2+2+2)/5$ & 1408/704/704 & 18/12 & 2.7699 / 3.7586 & 2.373 \\
W4 & Balanced & 2 & $4/(7+7)/4$ & 1152/592/592 & 22/15 & 2.7620 / 3.7507 & 2.363 \\
W4 & Balanced & 3 & $4/(6+6+6)/5$ & 896/496/496 & 27/15 & 2.7443 / 3.7330 & 2.366 \\
W4 & Balanced & 4 & $4/(6+6+6+6)/5$ & 768/432/432 & 33/15 & 2.7985 / 3.7872 & 2.367 \\
\bottomrule
\end{tabular*}
\end{table}
\subsubsection{Preserving depth with narrow branches}
\label{sec:narrow-depth-count}

To complement the branch-count study in Section~\ref{sec:multibranch-results},
Table~\ref{tab:narrow-depth-count} holds W1 depth at 12 with 6 blocks
per branch and 3 shared blocks on each side. Increasing branch count
requires narrower blocks at fixed compute and raises loss. Balanced
\mbt{}s instead trade some branch depth for greater width.

\begin{table}[!htb]
\captionsetup{skip=4pt}
\caption{W1 full-depth Narrow--Deep controls, reporting the best final loss
for each configuration. Compute is body / total PFLOP-days;
$L/D$ denotes physical block count / depth.}
\label{tab:narrow-depth-count}
\centering\fontsize{9}{11}\selectfont
\setlength{\tabcolsep}{4pt}
\renewcommand{\arraystretch}{1.00}
\begin{tabularx}{\linewidth}{Y c c r c}
\toprule
Model / allocation & Branch $d/m/f$ & $L/D$ & Loss $\downarrow$ & Body / Total \\
\midrule
Standard (12 blocks) & 1024/512/512 & 12/12 & 2.653 & 0.366514 / 0.620471 \\
Wide--Shallow $3/(3+3)/3$ & 1024/512/512 & 12/9 & 2.630 & 0.370569 / 0.624527 \\
\midrule
Narrow--Deep $3/(6+6)/3$ & 640/352/368 & 18/12 & 2.632 & 0.369463 / 0.623421 \\
Narrow--Deep $3/(6+6+6)/3$ & 512/272/272 & 24/12 & 2.650 & 0.369458 / 0.623416 \\
Narrow--Deep $3/(6+6+6+6)/3$ & 384/256/256 & 30/12 & 2.675 & 0.370701 / 0.624659 \\
\bottomrule
\end{tabularx}
\end{table}

\subsection{KV-cache sharing configurations and cache accounting}
\label{sec:shared-kv-details}

For Section~\ref{sec:inference-speedup}, Table~\ref{tab:shared-kv-all}
expands the W4 example in Table~\ref{tab:shared-kv-w4} to all scales,
combining configurations, final losses, cache savings, and training compute.
Within each configuration, only KV construction changes.

\begin{table}[!htb]
\captionsetup{skip=4pt}
\caption{Balanced \mbt{}s with Independent-KV and Joint-KV.
Losses are final validation losses; Joint-KV cache savings are relative to Standard.
Compute is body / total PFLOP-days for both variants.}
\label{tab:shared-kv-all}
\centering\fontsize{8}{9.5}\selectfont
\setlength{\tabcolsep}{3pt}\renewcommand{\arraystretch}{1.02}
\begin{tabular*}{\linewidth}{@{\extracolsep{\fill}}c c c r r r r@{}}
\toprule
\multicolumn{3}{c}{Architecture} & \multicolumn{2}{c}{Loss $\downarrow$} & \multicolumn{2}{c}{Cache and compute} \\
\cmidrule(lr){1-3}\cmidrule(lr){4-5}\cmidrule(l){6-7}
$K$ & Allocation & Branch $d/m/f$ & Independent & Joint & Joint KV saved & Body / Total \\
\midrule
\multicolumn{7}{l}{\textbf{W1}: Standard loss 2.653; Standard KV 24 KiB/token} \\
2 & $3/(5+5)/3$ & 768/368/384 & 2.618 & 2.620 & 18.75\% & 0.371744 / 0.625702 \\
\addlinespace[2pt]
3 & $3/(4+4+4)/3$ & 640/368/368 & 2.623 & 2.622 & 29.17\% & 0.374484 / 0.628442 \\
\addlinespace[2pt]
4 & $3/(3+3+3+3)/3$ & 640/368/368 & 2.626 & 2.633 & 34.38\% & 0.375752 / 0.629709 \\
\midrule
\multicolumn{7}{l}{\textbf{W2}: Standard loss 2.554; Standard KV 31.5 KiB/token} \\
2 & $3/(6+6)/3$ & 896/464/464 & 2.525 & 2.528 & 23.81\% & 0.783830 / 1.205417 \\
\addlinespace[2pt]
3 & $3/(5+5+5)/4$ & 768/352/352 & 2.529 & 2.537 & 26.19\% & 0.785588 / 1.207175 \\
\addlinespace[2pt]
4 & $3/(4+4+4+4)/4$ & 640/384/384 & 2.529 & 2.534 & 34.13\% & 0.778221 / 1.199809 \\
\midrule
\multicolumn{7}{l}{\textbf{W3}: Standard loss 2.471; Standard KV 40 KiB/token} \\
2 & $4/(6+6)/4$ & 1024/512/512 & 2.436 & 2.434 & 20.00\% & 1.530086 / 2.190659 \\
\addlinespace[2pt]
3 & $4/(5+5+5)/4$ & 896/432/448 & 2.446 & 2.451 & 28.13\% & 1.515967 / 2.176539 \\
\addlinespace[2pt]
4 & $4/(4+4+4+4)/4$ & 896/400/400 & 2.449 & 2.454 & 32.50\% & 1.514800 / 2.175372 \\
\midrule
\multicolumn{7}{l}{\textbf{W4}: Standard loss 2.392; Standard KV 49.5 KiB/token} \\
2 & $4/(7+7)/4$ & 1152/592/592 & 2.363 & 2.361 & 23.74\% & 2.762023 / 3.750740 \\
\addlinespace[2pt]
3 & $4/(6+6+6)/5$ & 896/496/496 & 2.366 & 2.366 & 28.79\% & 2.744269 / 3.732987 \\
\addlinespace[2pt]
4 & $4/(6+6+6+6)/5$ & 768/432/432 & 2.367 & 2.364 & 31.82\% & 2.798531 / 3.787248 \\
\bottomrule
\end{tabular*}
\end{table}

\textbf{Cache capacity.}
Let Standard have $L_0$ blocks of width $d_o$. An \mbt{} has $P$ \texttt{prelude}
blocks, $C$ \texttt{coda} blocks, and $K$ branches of $q$ blocks at width $d_b$.
With our 2:1 GQA ratio, BF16 keys and values occupy $2d$ bytes per token
per layer. Full-model capacities per token are
\begin{align*}
  M_{\mathrm{Standard}} &= 2L_0d_o,\\
  M_{\mathrm{Independent}} &= 2\bigl[(P+C)d_o+Kqd_b\bigr],\\
  M_{\mathrm{Shared}} &= 2\bigl[(P+C)d_o+qd_b\bigr].
\end{align*}
Savings are $1-M/M_{\mathrm{Standard}}$ at fixed batch size, context
length, and precision. Standard occupies 24, 31.5, 40, and
49.5 KiB/token for W1--W4. These capacities count stored keys and
values, excluding runtime overhead and cross-device replication.

\textbf{Training compute.}
Joint-KV preserves matrix FLOPs: projecting the concatenated
$K$ branch states once costs the same as $K$ independent projections,
and each branch still performs attention. Body and total costs in
Table~\ref{tab:shared-kv-all} follow Appendix~\ref{sec:evaluation-protocol};
total includes the tied LM head.

\textbf{Downstream evaluation.}
For Section~\ref{sec:inference-speedup},
Table~\ref{tab:joint-kv-downstream} reports the nine zero-shot tasks
for the final 2-, 3-, and 4-branch Balanced Joint-KV \mbt{}s in
Table~\ref{tab:shared-kv-all}. We use the same Standard checkpoints
and evaluation protocol as Table~\ref{tab:results} and
Appendix~\ref{sec:evaluation-protocol}.
\begin{table}[!htbp]
\centering
\caption{Zero-shot downstream performance of Balanced Joint-KV \mbt{}s and Standard.
Scores are percentages ($\uparrow$); the best displayed score within each group is bold.}
\label{tab:joint-kv-downstream}
\begingroup
\fontsize{9}{11}\selectfont
\setlength{\tabcolsep}{2.4pt}
\renewcommand{\arraystretch}{1.05}
\begin{tabular*}{\linewidth}{@{\extracolsep{\fill}}l l*{10}{r}@{}}
\toprule
Scale & Model & HS & ARC-E & ARC-C & OBQA & SciQ & WG & BoolQ & LMDA & PIQA & Average \\
\midrule
W1 & Standard & 45.15 & 57.37 & 31.06 & 34.40 & 75.20 & \textbf{54.85} & 56.36 & 37.42 & \textbf{69.80} & 51.29 \\
 & Joint-KV, $K=2$ & \textbf{47.65} & 56.40 & \textbf{31.66} & \textbf{36.80} & \textbf{79.60} & 54.30 & \textbf{60.00} & 38.29 & 69.70 & \textbf{52.71} \\
 & Joint-KV, $K=3$ & 47.33 & 57.74 & 31.06 & 34.20 & 76.90 & 53.75 & 56.64 & \textbf{38.91} & 69.53 & 51.78 \\
 & Joint-KV, $K=4$ & 47.07 & \textbf{58.63} & 31.31 & 35.00 & 75.80 & 53.20 & 58.17 & 36.29 & 68.55 & 51.56 \\
\midrule
W2 & Standard & 50.29 & 62.46 & 33.45 & 35.20 & 80.50 & 53.28 & 47.22 & 38.95 & 71.00 & 52.48 \\
 & Joint-KV, $K=2$ & \textbf{52.29} & 61.11 & \textbf{35.67} & 37.00 & \textbf{82.70} & 53.51 & 58.04 & 40.03 & 70.67 & 54.56 \\
 & Joint-KV, $K=3$ & 51.71 & 62.12 & 34.98 & 35.00 & 81.00 & \textbf{56.99} & \textbf{61.28} & 39.49 & 71.27 & 54.87 \\
 & Joint-KV, $K=4$ & 51.79 & \textbf{64.65} & 34.64 & \textbf{38.00} & 81.80 & 53.28 & 60.76 & \textbf{40.36} & \textbf{71.71} & \textbf{55.22} \\
\midrule
W3 & Standard & 53.30 & 63.47 & 35.58 & 37.60 & 84.50 & \textbf{57.85} & 57.55 & 40.56 & 72.36 & 55.86 \\
 & Joint-KV, $K=2$ & \textbf{55.65} & 65.24 & 36.35 & 38.00 & \textbf{85.00} & 57.54 & 56.48 & \textbf{44.15} & 72.14 & \textbf{56.73} \\
 & Joint-KV, $K=3$ & 54.76 & \textbf{65.78} & \textbf{37.20} & 38.80 & 83.40 & 54.93 & 58.99 & 43.20 & 71.82 & 56.54 \\
 & Joint-KV, $K=4$ & 54.54 & 63.97 & 36.09 & \textbf{39.40} & 82.20 & 55.72 & \textbf{61.77} & 42.11 & \textbf{72.69} & 56.50 \\
\midrule
W4 & Standard & 57.34 & 65.66 & 37.46 & 40.60 & 81.10 & 56.43 & 58.78 & 44.52 & 73.61 & 57.28 \\
 & Joint-KV, $K=2$ & \textbf{59.17} & \textbf{68.86} & \textbf{40.61} & 39.00 & \textbf{86.80} & 57.85 & 61.93 & 48.13 & \textbf{74.86} & \textbf{59.69} \\
 & Joint-KV, $K=3$ & 58.54 & 66.54 & 38.14 & \textbf{41.00} & 85.70 & \textbf{58.09} & \textbf{62.17} & 46.24 & 74.81 & 59.03 \\
 & Joint-KV, $K=4$ & 59.04 & 66.75 & 38.65 & 40.20 & 84.90 & 57.85 & 56.82 & \textbf{48.52} & 74.65 & 58.60 \\
\midrule
Mean & Standard & 51.52 & 62.24 & 34.39 & 36.95 & 80.33 & 55.60 & 54.98 & 40.36 & 71.69 & 54.23 \\
 & Joint-KV, $K=2$ & \textbf{53.69} & 62.90 & \textbf{36.07} & 37.70 & \textbf{83.53} & 55.80 & 59.11 & \textbf{42.65} & 71.84 & \textbf{55.92} \\
 & Joint-KV, $K=3$ & 53.09 & 63.05 & 35.35 & 37.25 & 81.75 & \textbf{55.94} & \textbf{59.77} & 41.96 & 71.86 & 55.56 \\
 & Joint-KV, $K=4$ & 53.11 & \textbf{63.50} & 35.17 & \textbf{38.15} & 81.17 & 55.01 & 59.38 & 41.82 & \textbf{71.90} & 55.47 \\
\bottomrule
\end{tabular*}
\endgroup
\par\vspace{3pt}
\begin{minipage}{\linewidth}\fontsize{8}{10}\selectfont
$K$: number of branches. Average weights the nine tasks equally; Mean averages each task over W1--W4.
Task definitions and metrics follow Appendix~\ref{sec:evaluation-protocol}; averages use unrounded scores.
\end{minipage}
\end{table}

\subsection{Width and depth controls}
\label{sec:width-depth-details}
For Section~\ref{sec:analysis-width-depth} and Figure~\ref{fig:width-depth-ablation},
Table~\ref{tab:width-depth-controls} expands the W1 example in
Table~\ref{tab:w1-width-depth-example} to all four scales. The Standard and two-branch
Wide--Shallow references are fixed while the three control families
change the allocation of width and depth.
\begin{table}[!htbp]
  \caption{All five families in Figure~\ref{fig:width-depth-ablation}.
  Widths are residual / MoE latent / expert FFN ($d/m/f$).
  Shared widths follow Table~\ref{tab:training-config}; all configurations
  retain the per-scale training-token budget.}
  \label{tab:width-depth-controls}
  \centering\fontsize{8}{10}\selectfont
  \setlength{\tabcolsep}{4pt}
  \begin{tabular}{l c c c c c}
    \toprule
    Family & Scale & Allocation & \shortstack{Physical blocks /\\depth}
      & \shortstack{Shared or all-layer\\$d/m/f$}
      & \shortstack{Core or branch\\$d/m/f$} \\
    \midrule
    Standard & W1 & $12$ & 12 / 12 & 1024/512/512 & --- \\
     & W2 & $14$ & 14 / 14 & 1152/576/576 & --- \\
     & W3 & $16$ & 16 / 16 & 1280/640/640 & --- \\
     & W4 & $18$ & 18 / 18 & 1408/704/704 & --- \\
    \addlinespace[3pt]
    Global wide--shallow & W1 & $9$ & 9 / 9 & 1280/640/512 & --- \\
    (sequential) & W2 & $10$ & 10 / 10 & 1408/704/704 & --- \\
     & W3 & $12$ & 12 / 12 & 1536/768/720 & --- \\
     & W4 & $13$ & 13 / 13 & 1792/896/704 & --- \\
    \addlinespace[3pt]
    Local wide-core & W1 & $3/3/3$ & 9 / 9 & 1024/512/512 & 1664/640/640 \\
    (sequential) & W2 & $3/4/3$ & 10 / 10 & 1152/576/576 & 1920/704/704 \\
     & W3 & $4/4/4$ & 12 / 12 & 1280/640/640 & 2176/768/768 \\
     & W4 & $4/5/4$ & 13 / 13 & 1408/704/704 & 2304/880/944 \\
    \addlinespace[3pt]
    Wide--Shallow \mbt{} & W1 & $3/(3+3)/3$ & 12 / 9 & 1024/512/512 & 1024/512/512 \\
    (two branches) & W2 & $3/(4+4)/3$ & 14 / 10 & 1152/576/576 & 1152/576/576 \\
     & W3 & $4/(4+4)/4$ & 16 / 12 & 1280/640/640 & 1280/640/640 \\
     & W4 & $4/(5+5)/4$ & 18 / 13 & 1408/704/704 & 1408/704/704 \\
    \addlinespace[3pt]
    Narrow--Deep \mbt{} & W1 & $3/(6+6)/3$ & 18 / 12 & 1024/512/512 & 640/320/416 \\
    (two branches) & W2 & $3/(8+8)/3$ & 22 / 14 & 1152/576/576 & 768/384/384 \\
     & W3 & $4/(8+8)/4$ & 24 / 16 & 1280/640/640 & 896/400/416 \\
     & W4 & $4/(10+10)/4$ & 28 / 18 & 1408/704/704 & 896/448/560 \\
    \bottomrule
  \end{tabular}
\end{table}

\subsection{Cache-matched Joint-KV and Independent-KV comparison}
\label{sec:joint-gqa-details}

For Section~\ref{sec:analysis-joint-gqa}, Table~\ref{tab:joint-gqa-ladder}
provides widths and body compute for the paired comparisons in
Table~\ref{tab:joint-gqa-setup}.
Within each pair, both models retain the same allocation, widths,
query heads, outer blocks, and fusion. Joint-KV constructs 2 shared
KV heads at each branch depth; Independent-KV constructs 1 private head per branch.
All heads have dimension 64, and caches use BF16. Outer widths follow Table~\ref{tab:training-config}; query and KV head
counts are listed in Table~\ref{tab:joint-gqa-ladder}.

\begin{table}[!htbp]
\centering
\caption{W1--W4 configurations for Section~\ref{sec:analysis-joint-gqa}. Cache size is identical within each pair. Body compute (PFLOP-days) and loss report Joint-KV / Independent-KV at the same training-token budget.}
\label{tab:joint-gqa-ladder}
\begingroup
\fontsize{8}{10}\selectfont
\setlength{\tabcolsep}{3pt}
\renewcommand{\arraystretch}{1.12}
\begin{tabular*}{\linewidth}{@{\extracolsep{\fill}}lcccccccc@{}}
\toprule
 & & & \multicolumn{3}{c}{Q/KV heads}
 & & & \\
\cmidrule(lr){4-6}
Scale & Allocation & Branch $d/m/f$ & Outer
 & Joint-KV & Independent-KV & \shortstack{KV/token\\(KiB)} & Body compute & Loss $\downarrow$ \\
\midrule
W1 & $3/(5+5)/3$ & $768/368/384$ & $16/2$
 & $16/2$ & $16/1$ & 5.5 & $0.3867/0.3848$ & $2.646/2.657$ \\
W2 & $3/(6+6)/3$ & $896/464/464$ & $18/2$
 & $16/2$ & $16/1$ & 6.0 & $0.7763/0.7723$ & $2.557/2.568$ \\
W3 & $4/(6+6)/4$ & $1024/512/512$ & $20/2$
 & $16/2$ & $16/1$ & 7.0 & $1.4499/1.4436$ & $2.487/2.499$ \\
W4 & $4/(7+7)/4$ & $1152/592/592$ & $24/3$
 & $18/2$ & $18/1$ & 9.5 & $2.6664/2.6551$ & $2.380/2.392$ \\
\bottomrule
\end{tabular*}
\endgroup
\end{table}

Both variants are trained from scratch on the same ordered data and
token budget at each scale. 
Both follow the protocol in Appendix~\ref{sec:training-setup}, including the LR candidates and batch size.
All reported losses are measured at the end of this token budget. 
Joint-KV's larger
KV projection adds less than 2.1M parameters and 0.6\% body compute;
Table~\ref{tab:joint-gqa-ladder} records each model's body compute.
\subsection{Fusion implementation}
\label{sec:fusion-details}

This subsection gives the implementation details for the fusion ablation
in Section~\ref{sec:analysis-fusion} and Table~\ref{tab:fusion-operators}.

\textbf{Parameterization.}
$h_1,h_2\in\mathbb{R}^d$ are
normalized branch outputs, with $d=1280$ and auxiliary width $r=128$.
The weights have shapes $W_i\in\mathbb{R}^{d\times d}$,
$A_i,B_i\in\mathbb{R}^{r\times d}$, and $P\in\mathbb{R}^{d\times r}$.
Both gates take $x=[W_1h_1;W_2h_2]$:
\[
  g=\sigma\!\left(G_o\operatorname{SiLU}(G_i x)\right),
  \qquad
  \beta=\operatorname{softmax}\!\left(G_o\operatorname{SiLU}(G_i x)\right).
\]
Here $G_i\in\mathbb{R}^{r\times2d}$; $G_o$ has $d$ rows for channel
competition and 2 rows for token softmax. Softmax operates over branches
at each token, without uniform-weight smoothing. All projections are bias-free.

\textbf{Initialization and computation.}
The learned branch RMSNorms start with unit gains and use $\epsilon=10^{-5}$.
The concat projection uses normal standard deviation $0.02/\sqrt{32}$;
auxiliary inner matrices use 0.02. Zero-initialized $P$ and $G_o$ make
corrections vanish and gates equal, recovering concat for the learned
correction and gating variants.
We retain the fused concat projection and add corrections to preserve
the identical initial BF16 output. Gates and correction accumulation
use FP32; inner projections use BF16. Auxiliary work adds 0.96--1.09\%
body compute. Other settings follow the W3 training recipe.

\textbf{Projection-free control.}
The normalized sum uses $z=(h_1+h_2)/\sqrt{2}$ without a fusion matrix.
It retains the two learned RMSNorm gains ($2d=2{,}560$ parameters), so only the aggregation, not the whole module, is parameter-free.
Removing the concat projection saves $2d^2=3{,}276{,}800$ parameters.
\section{Inference}
\label{sec:inference-appendix}

This appendix describes the inference engine of Section~\ref{sec:system_method}, the workloads and measurement protocol of Section~\ref{sec:inference-speedup}, the decode speedups of Joint-KV \mbt{}s at all four scales, and an ablation of the engine mechanisms.

\subsection{Engine implementation}
\label{sec:inference-engine}

\textbf{Shared infrastructure.}
The engine extends nano-vLLM~\citep{yu2025nanovllm}. Standard and \mbt{}s share the same infrastructure: a paged KV cache~\citep{kwon2023pagedattention}, FlashAttention-3~\citep{shah2024flashattention3} for attention, and FlashInfer~\citep{ye2025flashinfer} for sampling. Each decode step is captured once as a CUDA graph and replayed.

\textbf{Branch grouping.}
At each depth of the branch region, the engine stacks the weights of the $K$ branches into one tensor. It also stacks their hidden states into a $K\times T\times d_b$ tensor, where $T$ is the number of tokens in the step and $d_b$ is the branch width. Each operator of a block then runs once for all branches, including a single attention call (see head folding below). In the MoE layer, the expert indices selected by branch $i$ are offset into a concatenated expert table of all branches. A single MoE kernel call thus serves the tokens of every branch.

\textbf{Head folding.}
All models use two query heads per KV head. Head folding reshapes the queries of the $K$ branches from $[KT, 2h_{\mathrm{KV}}, d_h]$ to $[T, 2Kh_{\mathrm{KV}}, d_h]$. The $2K$ query heads that share one KV head then form one group, so one attention call reads the shared bank once for all branches.

\begin{figure}[!htbp]
  \centering
  \includegraphics[width=\linewidth]{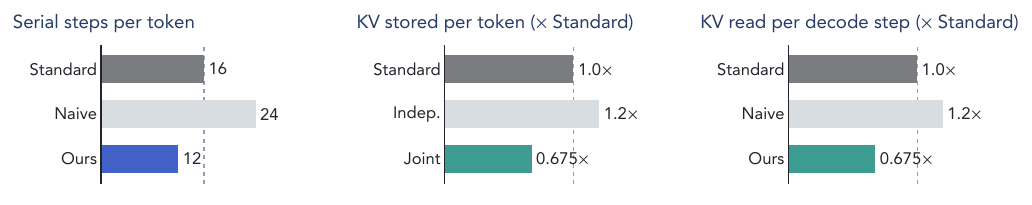}
  \caption{\textbf{Quantities changed by the serving mechanisms of Figure~\ref{fig:system-design}} for the 4-branch W3 Balanced \mbt{}, relative to Standard. All values are structural counts, not measurements. Left: serial block steps per token when the branches run one after another (Naive) and with branch grouping (Ours). Middle: KV cache stored per token. Right: KV read per decode step with Joint-KV, when each branch reads the shared bank separately (Naive) and with head folding (Ours).}
  \label{fig:system-counts}
\end{figure}

Figure~\ref{fig:system-counts} shows what each mechanism changes for the 4-branch W3 Balanced \mbt{}. Branch grouping cuts the serial steps per token from 24 to 12, against 16 for Standard. Joint-KV stores $0.675\times$ the KV cache of Standard, against $1.2\times$ for Independent-KV. Head folding cuts the KV read per decode step from $1.2\times$ to $0.675\times$ of Standard.

\textbf{Correctness.}
Before every timed run, we compare branch-grouped execution with head folding against sequential execution, which runs the branch blocks one at a time with one attention call per branch. On two prompts, the generated tokens, the logits, and the selected experts are bitwise identical in every reported setting. For W1--W4, all hidden states and KV caches are identical as well.

\subsection{Benchmark workloads and protocol}
\label{sec:inference-protocol}

\textbf{Prompts.}
All prompts are token sequences under the Llama~3 tokenizer~\citep{grattafiori2024llama3} used in training. They are stored in fixed prompt files, so every model receives identical token ids. The 4K-token prompts are the first 256 consecutive 4096-token windows of the tokenized validation split of FineWeb-Edu~\citep{penedo2024fineweb}, the corpus used in training. The prompts of 16K, 64K, and 128K tokens come from the official RULER generator~\citep{hsieh2024ruler}. We use seven of its tasks: five needle-in-a-haystack tasks (\texttt{niah\_single\_2}, \texttt{niah\_single\_3}, \texttt{niah\_multikey\_1}, \texttt{niah\_multivalue}, and \texttt{niah\_multiquery}) and two question-answering tasks (\texttt{qa\_1} and \texttt{qa\_2}). Each task has 40 samples per length for 16K and 64K and 20 for 128K, \textit{i.e.}, 280 and 140 prompts per length. Each sample is truncated from the left to exactly the target length, which keeps the question at the end.

\textbf{Requests.}
Requests run in rounds of $N$. In each round, $N$ requests start together. Each request is prefilled and then decodes exactly 1024 tokens, because an end-of-sequence token does not stop generation. The next round starts when all $N$ requests have finished. The rounds cycle through the first $4N$ prompts of the prompt file, with $N$ prompts per round. Sampling uses no top-$k$ truncation and a fixed seed.

\textbf{Measurement.} Timing starts after the model is loaded, the CUDA graphs are captured, and two warm-up rounds have run. Throughput is then measured in three windows, each lasting at least 60\,s and ending at the next round boundary. CUDA events split each round into prefill, which includes the first generated token, and decode. Decode throughput is the number of tokens generated in the decode phase over all $N$ requests, divided by the decode time. We report its mean over the three windows. Over the three windows, the decode throughput of every Joint-KV run in Figure~\ref{fig:decode-speedup-all} varies by at most 0.33\% (coefficient of variation). For 4K-, 16K-, and 64K-token inputs, all models of a scale run on one GPU in the order Standard, Joint-KV with $K{=}4$, $3$, and $2$, and Standard again. For 128K-token inputs, each Joint-KV model runs on its own GPU between two Standard runs. 

\subsection{Decode speedup at all scales}
\label{sec:inference-all-scales}

\begin{figure}[!htbp]
  \centering
  \input{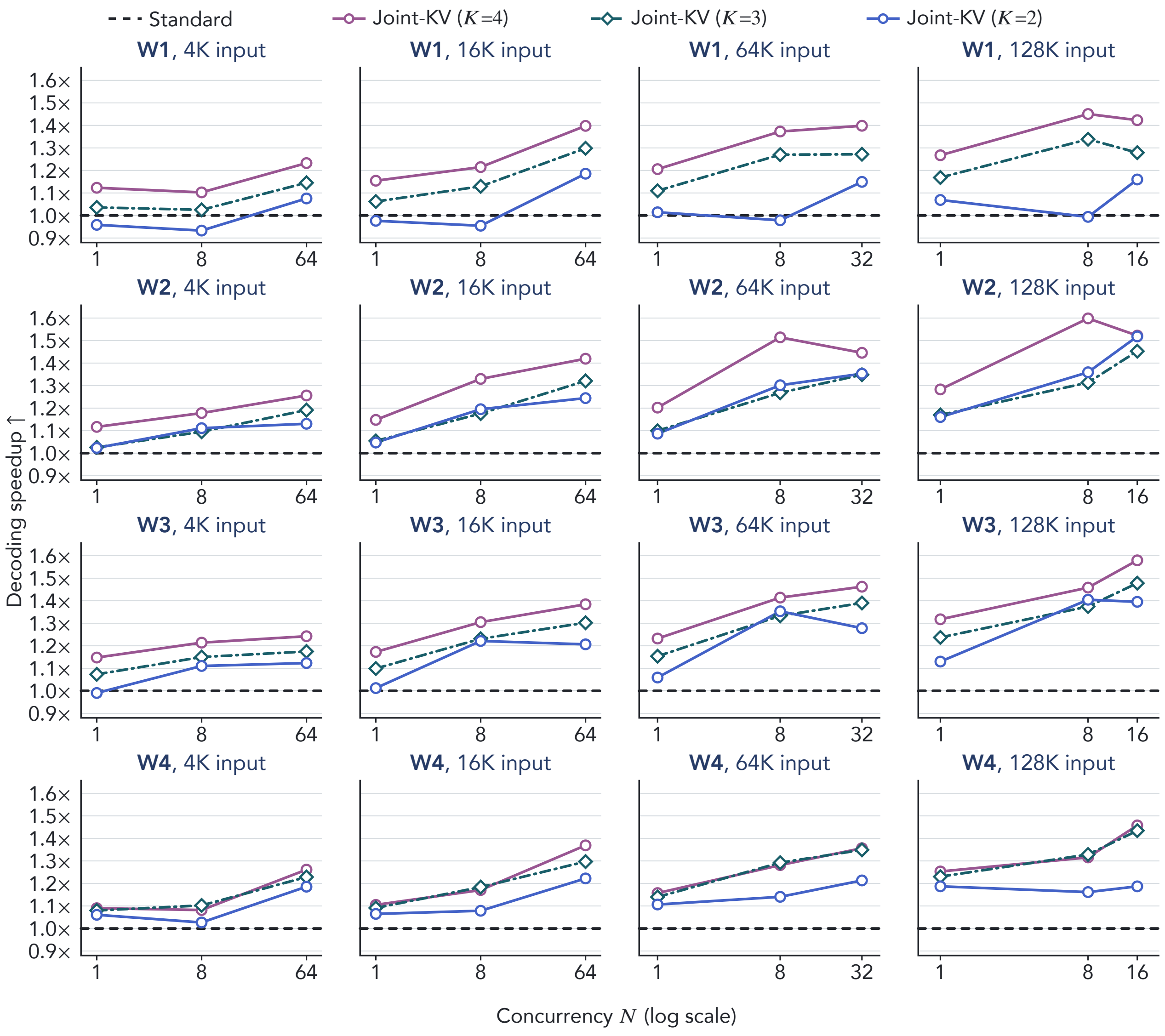}%
  \caption{\textbf{Decode speedup of Joint-KV \mbt{}s over Standard at all scales.} Decode throughput relative to Standard on one H200 GPU, as a function of the number of concurrent requests $N$, for 4K-, 16K-, 64K-, and 128K-token inputs that each generate 1K tokens. Rows are scales W1--W4, and all panels share the vertical axis.}
  \label{fig:decode-speedup-all}
\end{figure}

Figure~\ref{fig:decode-speedup-all} and Table~\ref{tab:decode-speedup-all} report the decode speedup of the Joint-KV \mbt{}s of Table~\ref{tab:shared-kv-all} at all four scales. Over the 48 settings of each model, four branches reach $1.08\times$ to $1.60\times$ and three branches $1.03\times$ to $1.48\times$. Two branches are slower than Standard in seven settings: six at W1 ($0.93\times$ to $0.99\times$) and one at W3 ($0.99\times$, 4K input, $N{=}1$).

\begin{table}[!htbp]
\caption{Depth, \textit{i.e.}, serial blocks per decode step, of Standard and of the Joint-KV \mbt{}s in Figure~\ref{fig:decode-speedup-all} (Eq.~\ref{eq:branch-path}).}
\label{tab:decode-depth}
\centering\fontsize{8}{9.5}\selectfont
\setlength{\tabcolsep}{6pt}
\begin{tabular}{l r r r r}
\toprule
Scale & Standard & $K{=}4$ & $K{=}3$ & $K{=}2$ \\
\midrule
W1 & 12 & 9 & 10 & 11 \\
W2 & 14 & 11 & 12 & 12 \\
W3 & 16 & 12 & 13 & 14 \\
W4 & 18 & 15 & 15 & 15 \\
\bottomrule
\end{tabular}
\end{table}

Per-kernel profiles of decode steps on one H200, with the same engine, explain these results. The profiled speedups agree with Table~\ref{tab:decode-speedup-all} within 2.5\%.

\textbf{Short inputs at low concurrency.} With a single 4K-token request, every \mbt{} reaches 0.84 to 0.91 of the depth ratio (Table~\ref{tab:decode-depth}). Most of the step scales with depth: excluding the operations that only \mbt{}s perform, a four-branch \mbt{} takes the depth-scaled step time of Standard within 4\%. The rest does not shrink with depth. The embedding, LM head, and sampling take 10 to 13\% of a Standard step. The \mbt{}-specific operations, namely the branch input map, the fusion, the bank construction, and head folding, add 100 to 160\,$\mu$s per step with four branches and 130 to 190\,$\mu$s with two. Branch counts with equal depth, such as three and four branches at W4, reach similar speedups. At W1, two branches remove only one block. This saves less than the \mbt{}-specific operations cost, so they are slower than Standard.

\textbf{Short inputs at higher concurrency.} Each layer routes a token to 8 of its 64 experts. More requests therefore make a step read more distinct experts, which accounts for most of the time added from $N{=}1$ to $N{=}8$ at 4K. This cost scales with the total expert parameters, which Balanced \mbt{}s keep close to Standard. With four branches, the ratio is 1.02 at W1, 0.89 at W3, and 1.00 at W4. The measured growth of expert time matches these ratios (1.05, 0.89, and 1.04). The speedup thus dips at W1 and W4. At $N{=}64$, nearly all experts are read, KV reads grow, and the speedup rises again.

\textbf{Long inputs at high concurrency.} KV reads dominate these steps, so the speedup approaches the bound $1/(1-s)$. For the four-branch W3 \mbt{}, this bound is $1.48$. With 32 concurrent 64K-token requests, the \mbt{} spends $0.668\times$ the attention time of Standard and reaches $1.46\times$. With 16 concurrent 128K-token requests, each step reads the same number of KV tokens as with 32 concurrent 64K-token requests. The \mbt{} then reaches $1.58\times$, which exceeds the bound. This happens because attention in the full-width layers runs slower at 128K and takes 27\% longer than at 64K. Attention over the shared banks does not slow down. All of the KV cache of Standard sits in full-width layers, but only 74\% of that of the \mbt{} does. The \mbt{} therefore reads KV at a higher average bandwidth (3.05 against 2.81\,TB/s), so its time saving exceeds its byte saving. This comparison is fair. Both models use the same engine, GPU, paged KV cache, and attention kernel and settings. The full-width layers of the \mbt{} slow down as much as those of Standard. The extra gain therefore comes from head folding.

\begin{table}[!htbp]
\caption{Decode speedup of Joint-KV \mbt{}s over Standard at all scales on one H200 GPU. The speedup is the decode throughput of an \mbt{} divided by that of Standard at the same input length and concurrency $N$. Drift (\%) is the change of the Standard throughput between the runs before and after the Joint-KV models. At 128K, each model has its own pair of Standard runs, and the table lists the largest of the three drifts. For each scale and row, the fastest model is in \textbf{bold} and the second fastest is \underline{underlined}.}
\label{tab:decode-speedup-all}
\centering\fontsize{8}{9.5}\selectfont
\setlength{\tabcolsep}{2.5pt}\renewcommand{\arraystretch}{1.05}
\begin{tabular*}{\linewidth}{@{\extracolsep{\fill}}l r r r r r r r r r r r r r r r r r@{}}
\toprule
 & & \multicolumn{4}{c}{W1} & \multicolumn{4}{c}{W2} & \multicolumn{4}{c}{W3} & \multicolumn{4}{c}{W4} \\
\cmidrule(lr){3-6}\cmidrule(lr){7-10}\cmidrule(lr){11-14}\cmidrule(lr){15-18}
Input & $N$ & $K{=}4$ & $K{=}3$ & $K{=}2$ & Drift & $K{=}4$ & $K{=}3$ & $K{=}2$ & Drift & $K{=}4$ & $K{=}3$ & $K{=}2$ & Drift & $K{=}4$ & $K{=}3$ & $K{=}2$ & Drift \\
\midrule
4K & 1 & \textbf{1.12} & \underline{1.04} & 0.96 & $-$4.2 & \textbf{1.12} & \underline{1.03} & 1.02 & $-$3.0 & \textbf{1.15} & \underline{1.07} & 0.99 & +0.5 & \textbf{1.09} & \underline{1.08} & 1.06 & $-$1.8 \\
 & 8 & \textbf{1.10} & \underline{1.03} & 0.93 & +6.2 & \textbf{1.18} & 1.09 & \underline{1.11} & +2.0 & \textbf{1.21} & \underline{1.15} & 1.11 & +1.6 & \underline{1.08} & \textbf{1.10} & 1.03 & +3.3 \\
 & 64 & \textbf{1.23} & \underline{1.15} & 1.08 & +0.3 & \textbf{1.26} & \underline{1.19} & 1.13 & +0.7 & \textbf{1.24} & \underline{1.17} & 1.12 & $-$0.1 & \textbf{1.26} & \underline{1.23} & 1.19 & $-$0.2 \\
\addlinespace[3pt]
16K & 1 & \textbf{1.15} & \underline{1.06} & 0.98 & +3.1 & \textbf{1.15} & \underline{1.05} & \underline{1.05} & +0.9 & \textbf{1.17} & \underline{1.10} & 1.01 & +2.8 & \textbf{1.10} & \underline{1.09} & 1.06 & +2.3 \\
 & 8 & \textbf{1.21} & \underline{1.13} & 0.95 & +0.6 & \textbf{1.33} & 1.18 & \underline{1.20} & +1.1 & \textbf{1.30} & \underline{1.23} & 1.22 & $-$0.5 & \underline{1.17} & \textbf{1.18} & 1.08 & +0.3 \\
 & 64 & \textbf{1.40} & \underline{1.30} & 1.19 & $-$0.2 & \textbf{1.42} & \underline{1.32} & 1.24 & $-$0.1 & \textbf{1.38} & \underline{1.30} & 1.21 & $-$0.3 & \textbf{1.37} & \underline{1.30} & 1.22 & $-$0.6 \\
\addlinespace[3pt]
64K & 1 & \textbf{1.21} & \underline{1.11} & 1.01 & +2.4 & \textbf{1.20} & \underline{1.10} & 1.09 & +0.7 & \textbf{1.23} & \underline{1.15} & 1.06 & +1.7 & \textbf{1.16} & \underline{1.14} & 1.11 & +2.0 \\
 & 8 & \textbf{1.37} & \underline{1.27} & 0.98 & $-$0.5 & \textbf{1.51} & 1.27 & \underline{1.30} & +0.3 & \textbf{1.41} & 1.33 & \underline{1.35} & $-$0.2 & \underline{1.28} & \textbf{1.29} & 1.14 & $-$0.6 \\
 & 32 & \textbf{1.40} & \underline{1.27} & 1.15 & +0.7 & \textbf{1.45} & \underline{1.35} & \underline{1.35} & +0.2 & \textbf{1.46} & \underline{1.39} & 1.28 & +0.6 & \textbf{1.36} & \underline{1.35} & 1.21 & +1.0 \\
\addlinespace[3pt]
128K & 1 & \textbf{1.27} & \underline{1.17} & 1.07 & +4.0 & \textbf{1.28} & \underline{1.17} & 1.16 & +1.1 & \textbf{1.32} & \underline{1.24} & 1.13 & $-$0.6 & \textbf{1.25} & \underline{1.23} & 1.19 & $-$0.9 \\
 & 8 & \textbf{1.45} & \underline{1.34} & 0.99 & +0.2 & \textbf{1.60} & 1.31 & \underline{1.36} & +0.7 & \textbf{1.46} & 1.37 & \underline{1.40} & $-$0.5 & \underline{1.32} & \textbf{1.33} & 1.16 & $-$0.8 \\
 & 16 & \textbf{1.42} & \underline{1.28} & 1.16 & +0.4 & \textbf{1.52} & \underline{1.45} & \textbf{1.52} & $-$0.2 & \textbf{1.58} & \underline{1.48} & 1.40 & $-$0.5 & \textbf{1.46} & \underline{1.43} & 1.19 & $-$0.2 \\
\bottomrule
\end{tabular*}
\end{table}

\subsection{Engine ablation}
\label{sec:engine-ablation}

\textbf{Setup.}
We measure how much each engine mechanism of Section~\ref{sec:system_method} contributes to the decode speedup. We fix the 4-branch W3 Joint-KV \mbt{} and compare three engines on one GPU, with 4K-, 16K-, and 64K-token inputs. The naive engine runs the branch blocks one at a time, each with its own attention call. The second engine adds branch grouping. The third engine also adds head folding and is the full engine.

\begin{figure}[!htbp]
  \centering
  \input{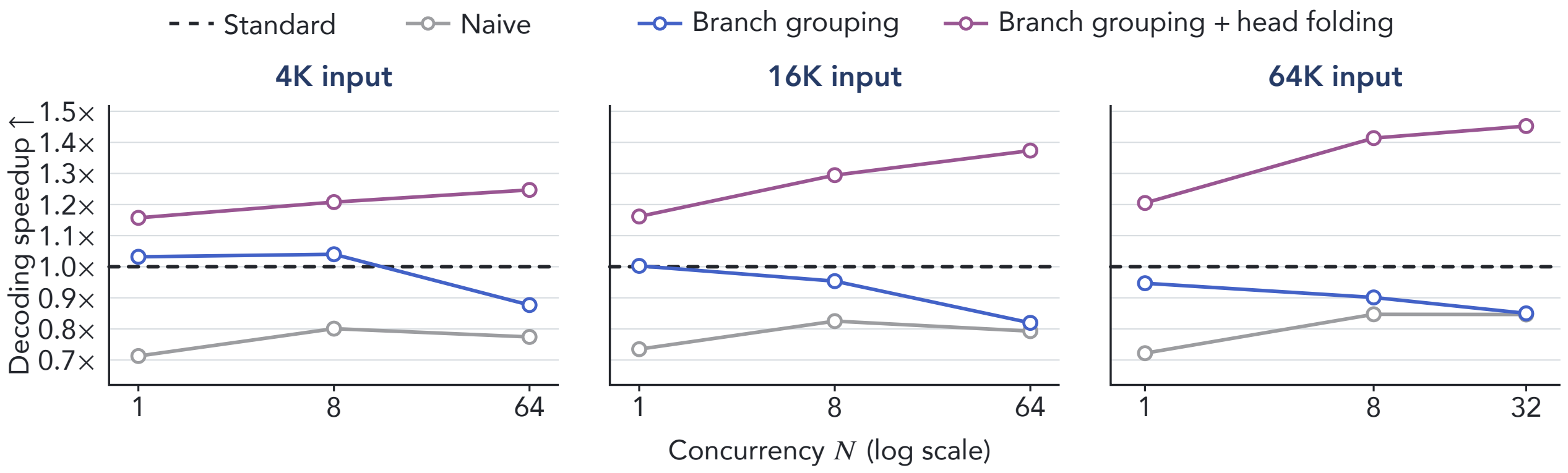}%
  \caption{\textbf{Contribution of each engine mechanism.} Decode speedup over Standard of the 4-branch W3 Joint-KV \mbt{} under three engines on one H200 GPU, for $N$ concurrent requests that each generate 1K tokens.}
  \label{fig:engine-ablation}
\end{figure}

\textbf{Results.}
Figure~\ref{fig:engine-ablation} shows the results. The naive engine is slower than Standard in every setting, because it runs more serial blocks and reads the shared KV cache once per branch. Branch grouping shortens the serial path, so it helps most with short inputs and low concurrency. Head folding removes the repeated KV reads, so its gain grows with the input length and concurrency. Only the two mechanisms together make the \mbt{} faster than Standard in every setting. Each mechanism reduces one of the two costs in Eq.~\ref{eq:decode-speedup}.

\end{document}